\documentclass[letterpaper, 10 pt, conference]{ieeeconf}  

\IEEEoverridecommandlockouts                              

\usepackage{times}
\usepackage{epsfig}
\usepackage{graphicx}
\usepackage{amsmath}
\usepackage{amssymb}

\usepackage{epsfig}
\usepackage{multirow}
\usepackage{tabularx}
\usepackage{adjustbox}
\usepackage{siunitx}
\usepackage{textcomp}
\usepackage{subfig}

\usepackage{stmaryrd} 
\usepackage{latexsym} 

\usepackage{url}

\usepackage{booktabs}

\newcommand{\figref}[1]{Fig.~\ref{#1}}

\newcommand{\tabref}[1]{Tab.~\ref{#1}}

\newcommand{\secref}[1]{Sec.~\ref{#1}}

\newcommand{\ie}{\textit{i.e.}}
\newcommand{\eg}{\textit{e.g.}}
\newcommand{\etal}{\textit{et al.}}

\usepackage{xcolor,colortbl}
\definecolor{Gray1}{gray}{0.85}
\definecolor{Gray2}{gray}{0.65}

\definecolor{darkgreen}{RGB}{0,127,0}
\definecolor{darkred}{RGB}{200,0,0}

\makeatletter
\let\NAT@parse\undefined
\makeatother
\usepackage[pagebackref=false,breaklinks=true,letterpaper=true,colorlinks,bookmarks=false]{hyperref}

\title{\LARGE \bf
Complementary Random Masking \\ for RGB-Thermal Semantic Segmentation
}

\author{Ukcheol Shin$^{1}$, Kyunghyun Lee$^{2}$, In So Kweon$^{2}$, Jean Oh$^{1}$
\thanks{$^{1}$U. Shin and J. Oh are with Robotics Institute, Carnegie Mellon University, Pittsburgh, Pennsylvania, 15217, United States
        {\tt\small \{ushin, hyaejino\}@andrew.cmu.edu}}%
\thanks{$^{2}$K. Lee and I. S. Kweon are with the School of Electrical Engineering, KAIST, Daejeon, 34141, Republic of Korea. 
{\tt\small \{kyunghyun.lee, iskweon77\}@kaist.ac.kr}}%
}

\begin{document}

\maketitle
\thispagestyle{empty}
\pagestyle{empty}

\begin{abstract}
RGB-thermal semantic segmentation is one potential solution to achieve reliable semantic scene understanding in adverse weather and lighting conditions.
However, the previous studies mostly focus on designing a multi-modal fusion module without consideration of the nature of multi-modality inputs.
Therefore, the networks easily become over-reliant on a single modality, making it difficult to learn complementary and meaningful representations for each modality.
This paper proposes 1) a complementary random masking strategy of RGB-T images and 2) self-distillation loss between clean and masked input modalities.
The proposed masking strategy prevents over-reliance on a single modality.
It also improves the accuracy and robustness of the neural network by forcing the network to segment and classify objects even when one modality is partially available.
Also, the proposed self-distillation loss encourages the network to extract complementary and meaningful representations from a single modality or complementary masked modalities.
We achieve state-of-the-art performance over three RGB-T semantic segmentation benchmarks.
Our source code is available at \url{https://github.com/UkcheolShin/CRM_RGBTSeg}.
\end{abstract}

\begin{figure}[t]
\begin{center}
{
\begin{tabular}{c@{\hskip 0.005\linewidth}c@{\hskip 0.005\linewidth}c}
\multicolumn{3}{c}{\includegraphics[width=0.96\linewidth]{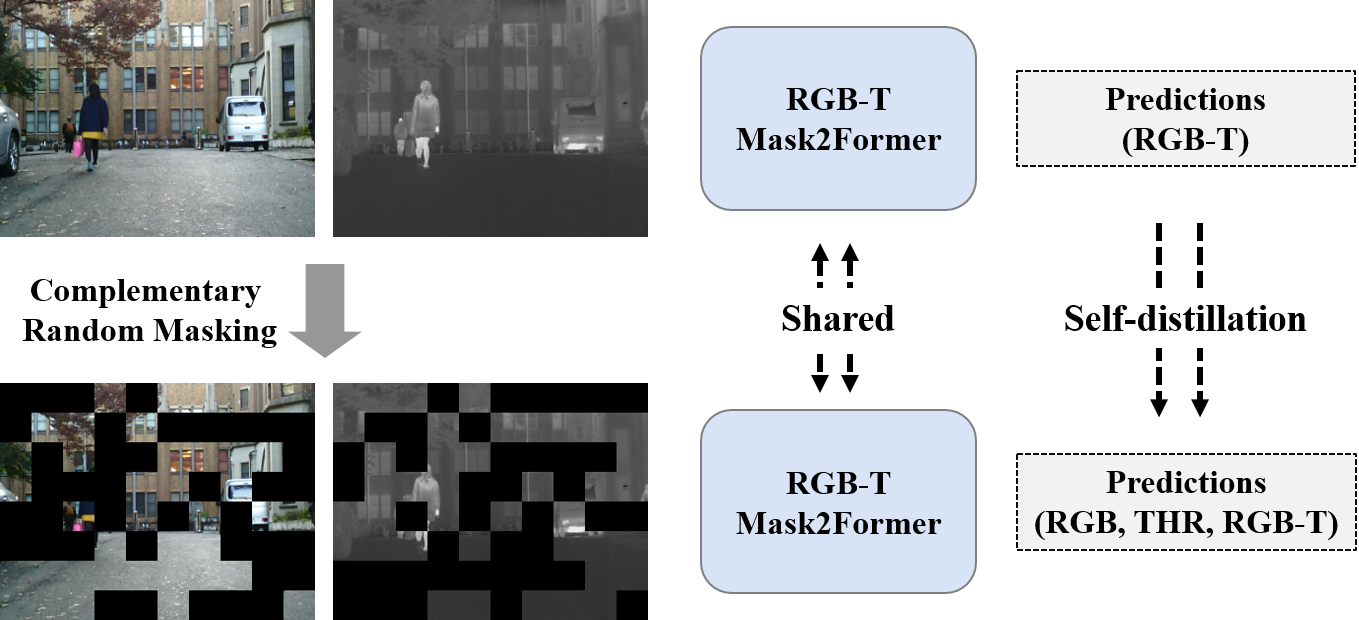}} \\ 
\multicolumn{3}{c}{{\footnotesize (a) Complementary random masking for RGB-T segmentation}} \\  
\includegraphics[width=0.32\linewidth]{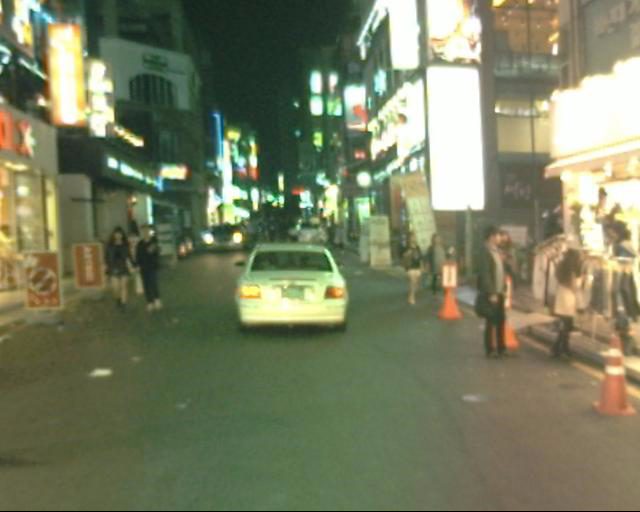} &
\includegraphics[width=0.32\linewidth]{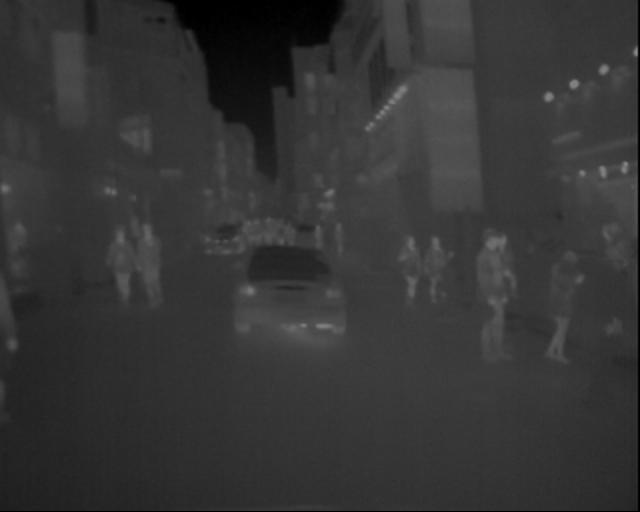} &
\includegraphics[width=0.32\linewidth]{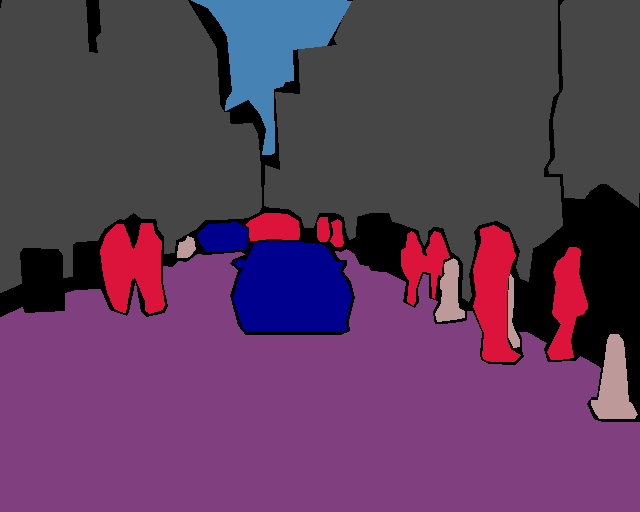} \\ 
{\footnotesize (b) RGB image } & {\footnotesize (c) Thermal image } & {\footnotesize (d) Ground Truth} \\ \includegraphics[width=0.32\linewidth]{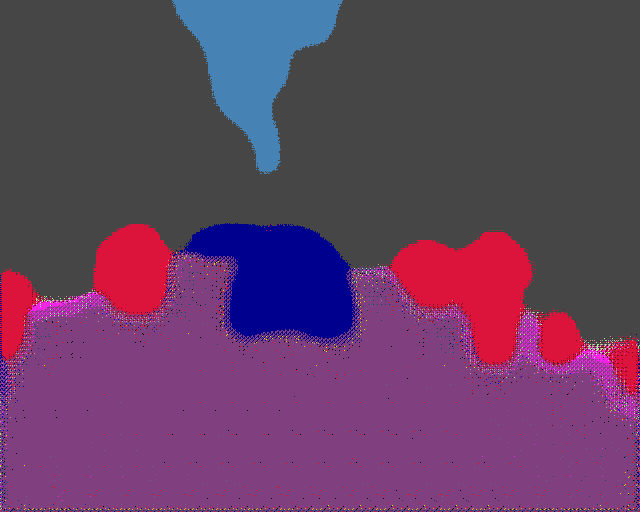} &
\includegraphics[width=0.32\linewidth]{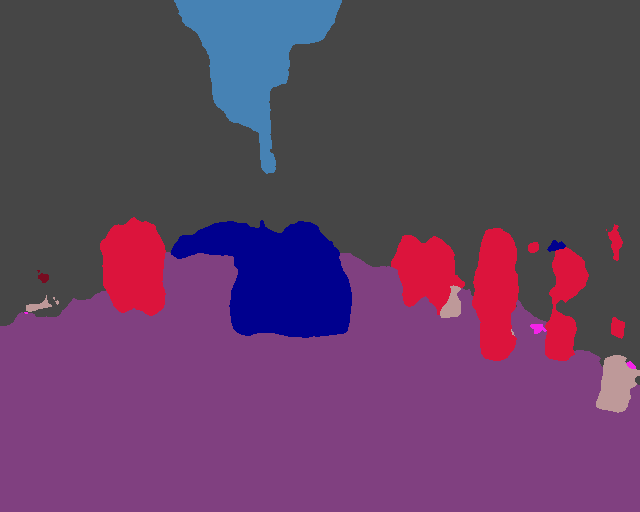} &
\includegraphics[width=0.32\linewidth]{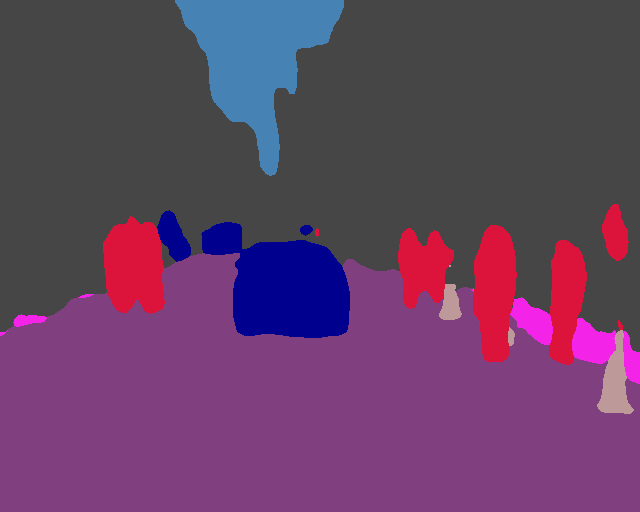} \\
{\footnotesize (e) RTFNet~\cite{sun2019rtfnet}} & {\footnotesize (f) CMXNet~\cite{liu2022cmx}} & {\footnotesize (g) Ours} \\
\end{tabular}
}
\end{center}
\caption{{\bf Complementary random masking for RGB-thermal semantic segmentation.}
Our proposed method aims to learn meaningful and complementary representations from RGB and thermal images by using complementary masking of RGB-T inputs and ensuring consistency between augmented and original inputs. 
The proposed method leads to robust and reliable segmentation results in day-light, low-light, and modality-dropped scenarios.
}
\label{fig:teaser}
\vspace{-0.1in}
\end{figure}

\section{Introduction}
\label{sec:intro}
Robust and reliable semantic scene understanding is a crucial fundamental ability for autonomous driving to ensure the safe and reliable operation of autonomous vehicles.
RGB-thermal semantic segmentation is one potential solution to achieve reliable semantic scene understanding in adverse weather and lighting conditions.
For example, in foggy or low-light conditions, the RGB camera may struggle to capture objects in the scene due to reduced visibility. 
In contrast, the thermal camera can still detect the heat signatures of objects. 
Combining the information from both modalities enables reliable and accurate semantic segmentation in adverse scenarios.
Therefore, it naturally led to recent active studies on semantic segmentation of RGB-thermal images~\cite{ha2017mfnet,sun2019rtfnet,shivakumar2019pst900,sun2020fuseseg,zhou2021gmnet,kim2021ms,liu2022cmx,zhao2023mitigating}.

\begin{figure}[t]
\begin{center}
{
\begin{tabular}{c@{\hskip 0.005\linewidth}c@{\hskip 0.005\linewidth}c}
\includegraphics[width=0.33\linewidth]{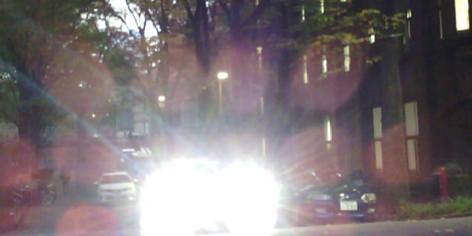} &
\includegraphics[width=0.33\linewidth]{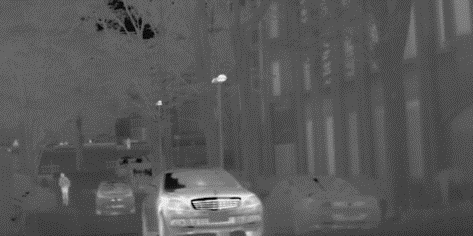} &
\includegraphics[width=0.33\linewidth]{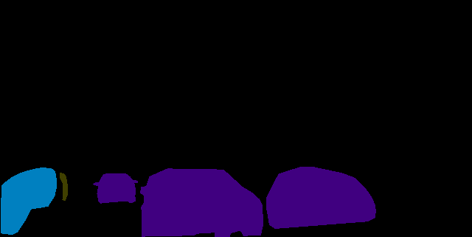} \\
{\footnotesize (a) RGB image} & {\footnotesize (b) Thermal image } & {\footnotesize (c) Ground Truth} \\
\includegraphics[width=0.33\linewidth]{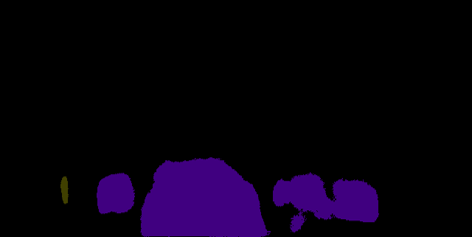} &
\includegraphics[width=0.33\linewidth]{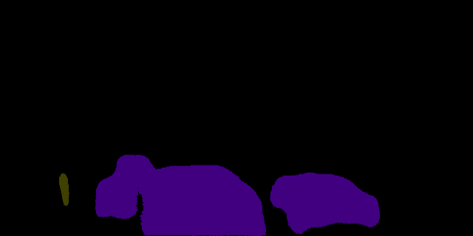} &
\includegraphics[width=0.33\linewidth]{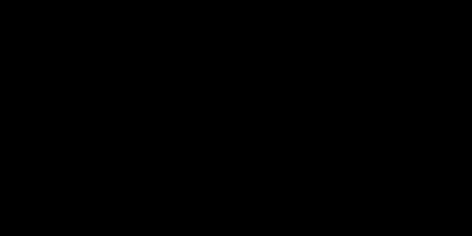} \\
{\footnotesize (d) RTFNet~\cite{sun2019rtfnet}} & {\footnotesize (e) RGB drop~\cite{sun2019rtfnet}} & {\footnotesize (f) THR drop~\cite{sun2019rtfnet}} \\
\includegraphics[width=0.33\linewidth]{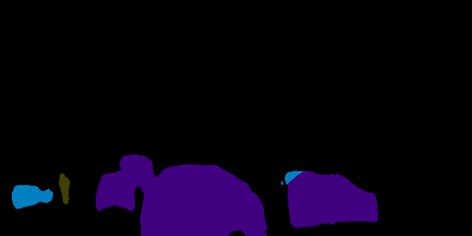} &
\includegraphics[width=0.33\linewidth]{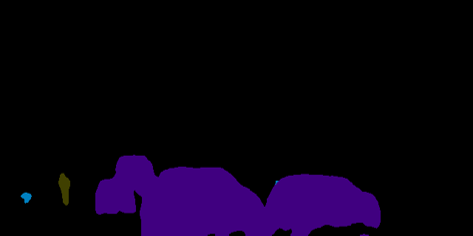} &
\includegraphics[width=0.33\linewidth]{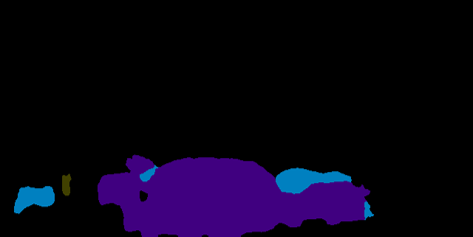} \\
{\footnotesize (g) Ours} & {\footnotesize (h) RGB drop (Ours)} & {\footnotesize (i) THR drop (Ours)} \\
\end{tabular}
}
\end{center}
\caption{{\bf Input modality dependency comparison of RGB-T semantic segmentation networks.} 
Common multi-modal fusion approaches often result in a sub-optimal solution, where the neural network becomes over-reliant on a single modality, as shown in (e) and (f).
On the other hand, our proposed method prevents the over-reliance issue (\ie, (h) and (i)). 
}
\label{fig:motiv}
\end{figure}

The primary research direction is designing a multi-modal fusion module~\cite{xu2021attention,zhang2021abmdrnet,deng2021feanet,liu2022cmx} that can effectively combine the information from both RGB and thermal modalities to improve the accuracy of semantic segmentation. 
However, without consideration of the nature of multi-modal inputs, the networks easily fall into a sub-optimal solution, where the network becomes over-reliant on a single modality, as shown in~\figref{fig:motiv} and~\tabref{tab:Motiv}.
In addition, this implies that the networks are susceptible to a wide range of fault cases, such as sensor disconnection, lens occlusion, and other input quality degeneration. 

In this paper, we focus on learning complementary and meaningful representations from both RGB and thermal modalities to prevent the over-reliance problem on a single modality and eventually improve the accuracy of the segmentation model.
For this purpose, our intuitive ideas are as follows and shown in~\figref{fig:teaser}:
1) We augment input RGB-T images with random masking to prevent the network from over-reliantly utilizing one modality for the semantic segmentation task. 
2) We enforce consistency between the prediction results of augmented and original images to encourage the network to extract meaningful representations even from partially occluded modalities or a single modality.

Our contributions can be summarized as follows:
\begin{itemize}
\item 
We propose a complementary random masking strategy that randomly masks one modality and masks the other modality in a complementary manner to improve model robustness and accuracy.
\item 
We propose a self-distillation loss between the prediction result from clean input modalities and the multiple prediction results from masked input modalities to learn complementary and non-local representations.
\item
Ours proposed method achieves state-of-the-art results over three RGB-T benchmark datasets (\ie, MFNet~\cite{ha2017mfnet}, PST900~\cite{shivakumar2019pst900}, and KP~\cite{hwang2015multispectral,kim2021ms} datasets).
\end{itemize}

\section{Related Work}
\label{sec:related}

\subsection{RGB-Thermal Semantic Segmentation}
Recently, thermal images have been widely adopted in various applications, such as detection~\cite{lee2022sequential,yang2022baanet}, tracking~\cite{li2019rgb,kang2022robust}, feature matching~\cite{lee2023edge, 9359356}, depth estimation~\cite{shin2021self,shin2022maximizing,shin2023self}, and SLAM~\cite{shin2019sparse,khattak2020keyframe}, to achieve high-level robustness against adverse weather and lighting conditions.

\begin{table}[t]
\caption{\textbf{Quantitative comparisons of RGB-T segmentation on MF dataset~\cite{ha2017mfnet} by input modality.}  
Previous RGB-T segmentation networks~\cite{sun2019rtfnet,liu2022cmx} show a critical vulnerability to modality drop and over-reliance on a single modality, which can hinder the learning of complementary representations from multiple modalities.}
\begin{center}
\resizebox{0.98\linewidth}{!}
{
    \def\arraystretch{1.2}
    \footnotesize
    \begin{tabular}{c|c|c|c|c|c}
    \toprule
    \multirow{2}{*}{Methods} & \textbf{RGB-T} & \multicolumn{2}{c|}{\textbf{RGB drop}} & \multicolumn{2}{c}{\textbf{THR drop}} \\ \cline{2-6}
    & mIoU $\uparrow$ & mIoU $\uparrow$ & Diff $\downarrow$ & mIoU $\uparrow$ & Diff $\downarrow$ \\
    \hline \hline
    RTFNet & 53.2 & 45.6 & \textcolor{red}{-7.6} & 10.5 & \textcolor{red}{-42.7} \\
    CMXNet & 58.0 & 44.7 & \textcolor{red}{-13.3} & 39.2 & \textcolor{red}{-18.8} \\
    Ours & \textbf{61.2} & \textbf{53.1} & \textcolor{red}{\textbf{-8.1}} & \textbf{52.7} & \textcolor{red}{\textbf{-8.5}} \\
    \bottomrule
    \end{tabular}
}
\end{center}
\label{tab:Motiv}
\end{table}

RGB-thermal semantic segmentation networks have also been proposed to overcome the limitation of RGB semantic segmentation networks~\cite{chen2017deeplab,he2017mask,wang2020deep,cheng2021maskformer,cheng2022masked} that are often vulnerable to extreme conditions, such as low-light, rainy, snowy, and sandy conditions.
Most previous RGB-T fusion networks focused on designing a multi-modal fusion module that can effectively combine the information from both modalities to improve the accuracy of semantic segmentation.
Specifically, they proposed various types of feature fusion modules, such as the na\"ive feature-level fusion~\cite{ha2017mfnet,sun2019rtfnet,shivakumar2019pst900}, multi-scale feature fusion~\cite{sun2019rtfnet,zhang2021abmdrnet,sun2020fuseseg,liu2022cmx}, and attention-weighted fusion~\cite{xu2021attention, zhang2021abmdrnet, deng2021feanet, liu2022cmx}.

However, if the nature of multi-modal inputs is not considered in the network training, the networks easily become over-reliant on a single modality.
This can hinder the network from learning complementary and meaningful representations for each modality, which is necessary to accurately and robustly segment objects.

\subsection{RGB-Thermal Knowledge Distillation} 
Several studies~\cite{vertens2020heatnet,kim2021ms,feng2023cekd} have investigated the potential of using knowledge distillation between RGB and thermal modalities to improve performance in various recognition applications.
Specifically, Heatnet~\cite{vertens2020heatnet} utilizes knowledge distillation from daytime prediction results to nighttime to improve the performance of the RGB-T semantic segmentation network. 
MS-UDA~\cite{kim2021ms} and CEKD~\cite{feng2023cekd} distill the knowledge of RGB-T segmentation network to thermal image segmentation network.
In contrast to these previous works, this paper specifically focuses on knowledge distillation between clean and masked images for RGB-T semantic segmentation tasks.

\begin{figure*}[t]
\begin{center} 
    \begin{tabular}{@{}c} 
    \includegraphics[width=0.98\linewidth]{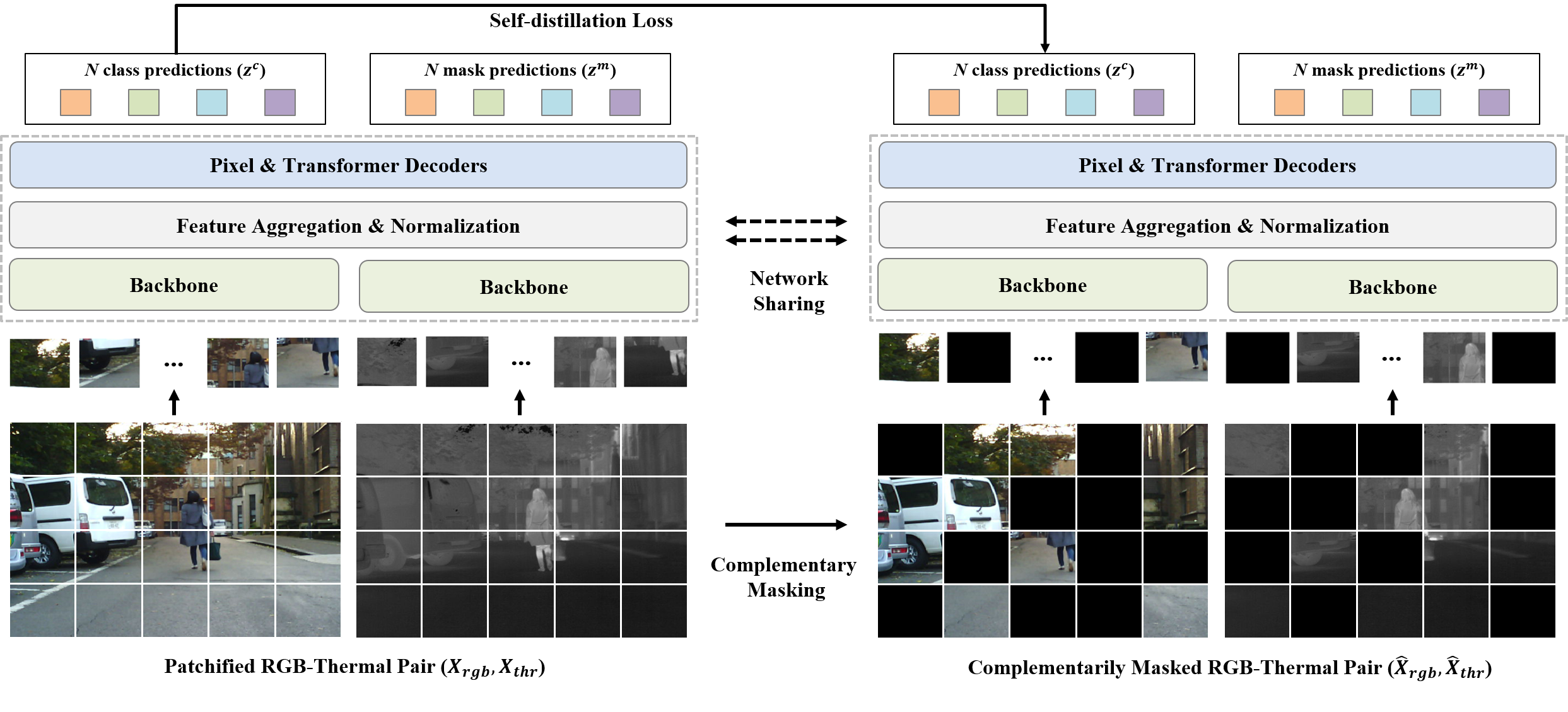}
    \end{tabular}
\end{center}
\vspace{-0.15in}
\caption{{\bf Overall pipeline of complementary masking and self-distillation for RGB-thermal semantic segmentation}. 
Our proposed training framework consists of complementary random masking and self-distillation loss.
We randomly mask the patchified RGB-thermal pair in a complementary manner that guarantees at least one modality is valid.  
After that, the network estimates each prediction results from clean and masked RGB-thermal pairs.
We enforce the network to predict the same class prediction results from the clean and masked RGB-thermal pairs.
The proposed method resolves the over-reliant problem of RGB-T semantic segmentation networks and encourages the network to extract complementary and meaningful representations for robust and accurate semantic segmentation performance from RGB-T images.}
\label{fig:overall_pipeline}
\end{figure*}

\section{Method} 
\label{sec:method}
\subsection{RGB-T Mask2Former}
\subsubsection{Preliminaries for Mask Classification}
Mask classification architecture~\cite{cheng2021maskformer,cheng2022masked} is a universal image segmentation network capable of semantic, instance, and panoptic segmentation. 
The network groups input pixels into $N$ segments by estimating $N$ binary masks and $N$ class labels.
The network consists of three main modules: a \textit{backbone} that extracts low-resolution features from an image, a \textit{pixel decoder} that gradually upsamples these features to generate high-resolution per-pixel embeddings, and a \textit{transformer decoder} that estimates object queries based on the image features. 
The class prediction is estimated via MLP layers with the object queries.
The binary mask predictions are obtained by decoding the per-pixel embeddings with object queries.
Please refer to these papers~\cite{cheng2021maskformer,cheng2022masked} for details. 

\subsubsection{Mask2Former for RGB-T images}
We adopted Mask2Former~\cite{cheng2022masked} for semantic segmentation as our baseline model and modified the model to take RGB and thermal images, as shown in~\figref{fig:overall_pipeline}.
More specifically, we assigned a modality-wise backbone for each modality. 
After extracting modality-wise image features from RGB and thermal images, we aggregate the features with a simple winner-take-all strategy via \textit{max} operation.
This aggregation finds the most prominent feature from the RGB and thermal features across channel dimensions.
We then normalized the aggregated feature map.
At this stage, we can directly forward one modality feature to the decoder without aggregating the multi-modal features.
After that, the aggregated multi-modal feature or single-modal feature is delivered to the pixel and transformer decoder to estimate $N$ class and mask predictions.
The final semantic segmentation mask can be obtained with a simple matrix multiplication of class and mask predictions.

\subsection{Complementary Random Masking}
Recently, masking strategy has been widely utilized in various language~\cite{devlin2018bert, liu2019roberta, brown2020language} and visual applications~\cite{bao2021beit,he2022masked,wei2022masked,tong2022videomae,xie2022simmim,pang2022masked} to learn meaningful representation.
Especially, the image masking strategy is used to pre-train a large capacity backbone model to learn general representation for various downstream tasks, such as recognition~\cite{he2022masked,wei2022masked}, video applications~\cite{tong2022videomae}, 3D application~\cite{pang2022masked}.
Differing from these works focusing on learning general representation, we utilize a masking strategy to overcome the over-reliant problem of the RGB-T semantic segmentation task and to learn complementary and robust representation for each modality.

More specifically, as shown in~\figref{fig:motiv} and~\tabref{tab:Motiv}, common RGB-T segmentation networks easily rely on a single modality. 
Therefore, the network rarely extracts meaningful representation from the other modality to segment and classify objects.
This makes the network vulnerable to a wide range of fault cases, such as sensor disconnection, lens occlusion, and other input quality degeneration. 
Also, it loses the chance to learn complementary and useful representations for each modality to segment and classify objects. 
Therefore, we push the network in a situation where one modality is partially unavailable but allows the missing information can be complemented by the other modality. 
For this purpose, we propose a complementary random masking method for RGB-T semantic segmentation.

\textbf{Complementary Patch Masking.} We use Swin-Transformer~\cite{liu2021swin} as our backbone model.
Therefore, each modality image is patchified into a set of non-overlapped small patches, which will be fed into the transformer model for processing.
Here, we randomly mask out the patches of one modality and replace the masked patches with learnable mask token vectors by following the convention of the token masking~\cite{devlin2018bert,bao2021beit,xie2022simmim}.
The other modality's patches are masked out in the same manner by using the complementary mask.
The complementary random masking process is defined as follows:
\begin{equation}
\begin{aligned}
\hat{X}_{rgb} &= M*X_{rgb} + \hat{M}*L_{rgb}, \\
\hat{X}_{thr} &= \hat{M}*X_{thr} + M*L_{thr},
\label{eqn:comp_rand_mask}
\end{aligned}
\end{equation}
where $X_{input}$ is tokenized input image, $M$ is random mask, $\hat{M}$ is its complementary mask, defined as $1-M$, and $L_{input}$ is learnable token vector.

\subsection{Self-distillation Loss}
The proposed self-distillation loss consists of two losses $L_{SDC}$ and $L_{SDN}$.
The former loss $L_{SDC}$ aims to make the network learn to extract complement representation when one modality information is partially unavailable.
Therefore, we enforce the network to predict the same class prediction results from clean (\ie, $X_{rgb}$ and $X_{thr}$) and masked (\ie, $\hat{X}_{rgb}$ and $\hat{X}_{thr}$) RGB-thermal pairs. 
The proposed self-distillation loss for complementary representations is defined as follows:
\begin{equation}
L_{SDC} = L_{1}(z^{c}(X_{rgb},X_{thr}), z^{c}(\hat{X}_{rgb},\hat{X}_{thr})), \\ 
\label{eqn:self_dist_loss_1}
\end{equation}
where $z^{c}(\cdot)$ is class prediction logit estimated from given tokenized inputs and $L_{1}(\cdot)$ is mean absolute error function.

The latter loss $L_{SDN}$ aims to make the network extract robust representations from a partially masked single modality based on their non-local context rather than local features. 
For this purpose, we further enforce the class prediction consistency between the clean RGB-T pair and single masked modality (\ie, $\hat{X}_{rgb}$ or $\hat{X}_{thr}$).
The proposed self-distillation loss for non-local representations is defined as follows:
\begin{equation}
\begin{aligned}
L_{SDN} &= L_{1}(z^{c}(X_{rgb},X_{thr}), z^{c}(\hat{X}_{rgb})), \\
       &+ L_{1}(z^{c}(X_{rgb},X_{thr}), z^{c}(\hat{X}_{thr})) 
\label{eqn:self_dist_loss_2}
\end{aligned}
\end{equation}

\subsection{Supervised Loss}
We utilize the same supervised loss function $L_{sup}$, which consists of binary mask loss $L_{mask}$ and classification loss $L_{cls}$, used in Mask2Former~\cite{cheng2022masked}. 
The supervised loss is defined as follows:
\begin{equation}
L_{sup}=L_{mask}+\lambda_{cls}L_{cls}, \\
\label{eqn:self_dist_loss_3}
\end{equation}
where the mask loss $L_{mask}$ is a combination of binary cross-entropy loss and dice loss~\cite{milletari2016v}, defined as $L_{mask}=\lambda_{ce}L_{ce}+\lambda_{dice}L_{dice}$. 

\textbf{Modality-wise Supervision.} The current network architecture can estimate three types of prediction results according to the given input modalities (\ie, RGB image, thermal image, and RGB-thermal pair).
We also empirically found that the supervised loss for each prediction result of multiple input modalities can perform better than a single supervised loss for RGB-thermal pair.
The modality-wise supervised loss is defined as follows:
\begin{equation}
\begin{aligned}
L_{MWS} &= L_{sup}(z_{gt},z(X_{rgb},X_{thr})), \\ 
            &+ L_{sup}(z_{gt},z(X_{rgb})) + L_{sup}(z_{gt},z(X_{thr})),
\label{eqn:modality_wise_sup}
\end{aligned}
\end{equation}
where $z_{gt}$ is ground truth class $z^c$ and binary mask $z^m$, $z$ is class and mask prediction from the given input modalities (RGB, thermal, or RGB-thermal). For the masked modalities, this loss only uses the first term (\ie, RGB-T). 
The total loss is defined as follows: 
\begin{equation}
L_{total} = L_{MWS} + L_{SDC} + L_{SDN}\\
\label{eqn:total_loss}
\end{equation}


\section{Experiments}
\label{sec:experiments}

\begin{table*}[t]
\caption{\textbf{Quantitative comparisons for semantic segmentation of RGB-T images on MF~\cite{ha2017mfnet}, PST900~\cite{shivakumar2019pst900}, and KP~\cite{hwang2015multispectral} datasets.} 
We compared our proposed method with the previous state-of-the-art RGB-T semantic segmentation networks on MF, PST900, and KP benchmarks.
Our proposed method demonstrates outperformed performance in all benchmark datasets. 
The best and second best performance in each block is highlighted in \textbf{bold} and \underline{underline}, respectively.}
\vspace{-0.1in}
\begin{center}

\subfloat[Quantitative comparison on MF day-night evaluation set~\cite{ha2017mfnet} (9 classes)] 
{
    \resizebox{0.90\linewidth}{!}
    {
        \def\arraystretch{0.9}
        \footnotesize
        \begin{tabular}{cccccccccc|c}
        \toprule
        Method & Unlabeled & Car & Person & Bike & Curve & Car Stop & Guardrail & Color Cone & Bump & {\bf mIoU}\\ 
        \hline \hline
        MFNet~\cite{ha2017mfnet}   	        & 96.9 & 65.9 & 58.9 & 42.9 & 29.9 & 9.9 & 0.0 & 25.2 & 27.7 & 39.7 \\
        PSTNet~\cite{shivakumar2019pst900} 	& 97.0 & 76.8 & 52.6 & 55.3 & 29.6 & 25.1 & \underline{15.1} & 39.4 & 45.0 & 48.4 \\
        RTFNet~\cite{sun2019rtfnet}  	    & \underline{98.5} & 87.4 & 70.3 & 62.7 & 45.3 & 29.8 & 0.0 & 29.1 & 55.7 & 53.2 \\
        FuseSeg~\cite{sun2020fuseseg}  	    & 97.6 & 87.9 & 71.7 & 64.6 & 44.8 & 22.7 & 6.4 & 46.9 & 47.9 & 54.5 \\
        AFNet~\cite{xu2021attention}  		& 98.0 & 86.0 & 67.4 & 62.0 & 43.0 & 28.9 & 4.6 & 44.9 & 56.6 & 54.6 \\
        ABMDRNet~\cite{zhang2021abmdrnet}  	& \textbf{98.6} & 84.8 & 69.6 & 60.3 & 45.1 & 33.1 & 5.1 & 47.4 & 50.0 & 54.8 \\
        FEANet~\cite{deng2021feanet}  	    & 98.3 & 87.8 & 71.1 & 61.1 & 46.5 & 22.1 & 6.6 & \underline{55.3} & 48.9 & 55.3 \\
        GMNet~\cite{zhou2021gmnet}  		& 97.5 & 86.5 & 73.1 & 61.7 & 44.0 & 42.3 & 14.5 & 48.7 & 47.4 & 57.3 \\
        CMX~\cite{liu2022cmx}               & 98.3 & \underline{90.1} & \underline{75.2} & 64.5 & \textbf{50.2} & 35.3 & 8.5 & 54.2 & \textbf{60.6} & 59.7 \\
        \hline
        Ours (Swin-T) & 98.2 & 90.0 & 73.1 & 63.7 & 47.9 & 40.7 & 9.9 & 54.4 & 54.2 & 59.1 \\
        Ours (Swin-S) & 98.4 & \textbf{90.6} & \textbf{75.5} & \textbf{67.2} & \underline{48.3} & \underline{43.4} & 11.8 & \textbf{56.8} & \underline{59.3} & \underline{61.2} \\
        Ours (Swin-B) & 98.2 & 90.0 & 75.1 & \underline{67.0} & 45.2 & \textbf{49.7} & \textbf{18.4} & 54.2 & 54.4 & \textbf{61.4} \\
        \bottomrule
        \end{tabular}
        \label{tab:MF_benchmark}
    }
} 

\quad
\subfloat[Quantitative comparison on PST900 evaluation set~\cite{shivakumar2019pst900} (5 classes)] 
{
    \resizebox{0.75\linewidth}{!}
    {
        \def\arraystretch{0.85}
        \footnotesize
        \begin{tabular}{cccccc|c}
        \toprule
        Method & Background & Fire-Extinguisher & Backpack & Hand-Drill & Survivor & {\bf mIoU}\\ 
        \hline \hline
        MFNet~\cite{ha2017mfnet}            & 98.6 & 60.4 & 64.3 & 41.1 & 20.7 & 57.0 \\ 
        RTFNet-152~\cite{sun2019rtfnet}     & 98.9 & 52.0 & 75.3 & 25.4 & 36.4 & 57.6 \\ 
        PSTNet~\cite{shivakumar2019pst900}  & 98.9 & 70.1 & 69.2 & 53.6 & 50.0 & 68.4 \\ 
        ABMDRNet~\cite{zhang2021abmdrnet}  	& 99.0 & 66.2 & 67.9 & 61.5 & 62.0 & 71.3 \\
        GMNet~\cite{zhou2021gmnet}          & 99.4 & 73.8 & 83.8 & 85.2 & 78.4 & 84.1 \\ 
        \hline
        Ours (Swin-T) & 99.5 & \underline{79.1} & 86.0 & 86.2 & 78.7 & 85.9 \\
        Ours (Swin-S) & \underline{99.6} & 76.2 & \textbf{91.0} & \underline{87.2} & \underline{80.8} & \underline{86.9} \\
        Ours (Swin-B) & \textbf{99.6} & \textbf{79.5} & \underline{89.6} & \textbf{89.0} & \textbf{82.2} & \textbf{88.0} \\ 
        \bottomrule
        \end{tabular}
        \label{tab:PST_benchmark}
    }
} 

\quad
\subfloat[Quantitative comparison on KP day-night evaluation set~\cite{hwang2015multispectral,kim2021ms} (19 classes)]
{
    \resizebox{0.90\linewidth}{!}
    {
        \def\arraystretch{1.2}
        \footnotesize
        \begin{tabular}{cccccccccccccccccccc|c} 
        \toprule
        Method&\rotatebox{90}{Road} & \rotatebox{90}{Sidewalk} & \rotatebox{90}{Building} & \rotatebox{90}{Wall} & \rotatebox{90}{Fence} & \rotatebox{90}{Pole} & \rotatebox{90}{Traffic light} & \rotatebox{90}{Traffic sign} & \rotatebox{90}{Vegetation} & \rotatebox{90}{Terrain} & \rotatebox{90}{Sky}  & \rotatebox{90}{Person} & \rotatebox{90}{Rider} & \rotatebox{90}{Car} & \rotatebox{90}{Truck} & \rotatebox{90}{Bus} & \rotatebox{90}{Train} & \rotatebox{90}{Motorcycle} & \rotatebox{90}{Bicycle} & \bf{mIoU} \\ 
        \hline \hline
        MFNet~\cite{ha2017mfnet}            & 93.5 & 23.6 & 75.1 & 0.0 & 0.1 & 9.1 & 0.0 & 0.0 & 69.3 & 0.2 & 90.4 & 24.0 & 0.0 & 69.6 & 0.3 & 0.3 & 0.0 & 0.0 & 0.6 & 24.0 \\ 
        RTFNet~\cite{sun2019rtfnet}         & 94.6 & 39.4 & 86.6 & 0.0 & 0.6 & 0.0 & 0.0 & 0.0 & 81.7 & 3.7 & 92.8 & 58.4 & 0.0 & 87.7 & 0.0 & 0.0 & 0.0 & 0.0 & 0.5 & 28.7 \\ 
        CMX~\cite{liu2022cmx}     & 97.7 & 53.8 & 90.2 & 0.0 & 47.1 & 46.2 & 10.9 & 45.1 & 87.2 & 34.3 & 93.5 & 74.5 & 0.0 & 91.6 & 0.0 & 59.7 & 0.0 & 46.1 & 0.2 & 46.2 \\
        \hline
        Ours (Swin-T) & 98.8 & 56.4 & 89.0 & 0.0 & 62.3 & 54.1 & 31.2 & \underline{31.2} & 84.3 & 23.2 & \underline{94.4} & \underline{83.6} & 0.0 & 94.7 & 0.0 & 77.7 & 0.0 & \underline{51.4} & 40.7 & 51.2 \\
        Ours (Swin-S) & \underline{98.8} & \underline{60.7} & \textbf{92.1} & 0.0 & \textbf{60.4} & \textbf{55.1} & 31.1 & 53.2 & \underline{89.1} & \textbf{27.7} & \textbf{95.0} & 81.4 & \textbf{17.7} & \underline{95.2} & \textbf{1.1} & \textbf{83.3} & 0.0 & 49.9 & \underline{42.3} & \underline{54.4} \\
        Ours (Swin-B) & \textbf{99.0} & \textbf{61.9} & 91.8 & 0.0 & 58.7 & 50.6 & \textbf{39.2} & \textbf{55.3} & \textbf{89.2} & \underline{23.2} & 94.3 & \textbf{85.2} & \underline{2.9} & \textbf{95.3} & 0.0 & \underline{80.5} & 0.0 & \textbf{66.2} & \textbf{54.6} & \textbf{55.2} \\
        \bottomrule
        \end{tabular}
        \label{tab:KP_benchmark}
    }
}
\end{center}
\vspace{-0.35in}
\label{tab:Exp_segmentation_total}
\end{table*}

\begin{figure*}[t]
\begin{center}
{
\begin{tabular}{c@{\hskip 0.005\linewidth}c@{\hskip 0.005\linewidth}c@{\hskip 0.005\linewidth}c@{\hskip 0.005\linewidth}c@{\hskip 0.005\linewidth}c}
\multicolumn{6}{c}{\includegraphics[width=0.98\linewidth]{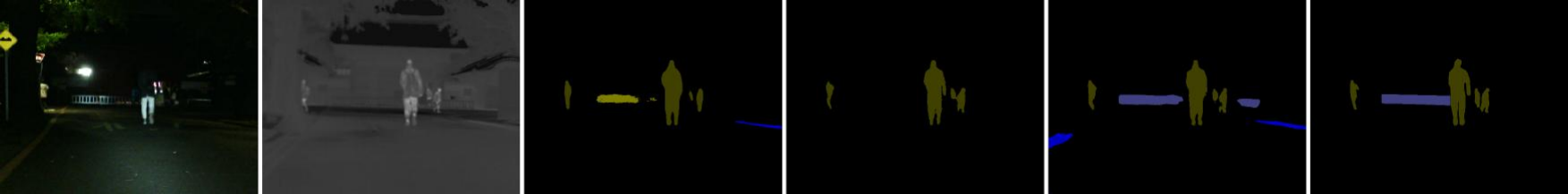}} \vspace{-0.02in} \\
\multicolumn{6}{c}{\includegraphics[width=0.98\linewidth]{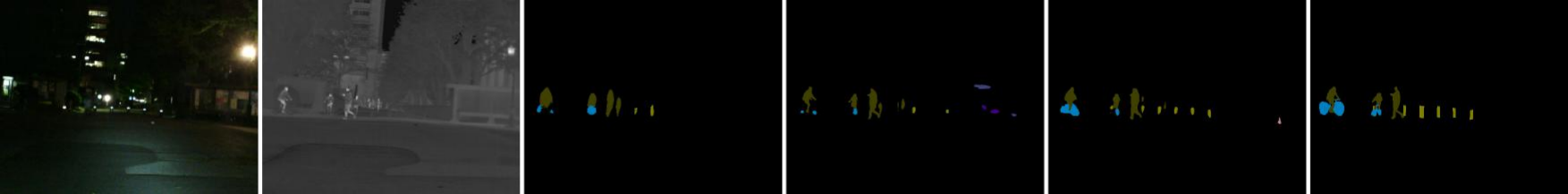}} \vspace{-0.02in} \\
{\footnotesize (a) RGB image} & {\footnotesize (b) THR image} & {\footnotesize (c) RTFNet~\cite{sun2019rtfnet} } & {\footnotesize (d) CMXNet~\cite{liu2022cmx}} &  {\footnotesize (e) Ours (Swin-B)} &  {\footnotesize (e) GT}  \\ 
\multicolumn{6}{c}{\includegraphics[width=0.98\linewidth]{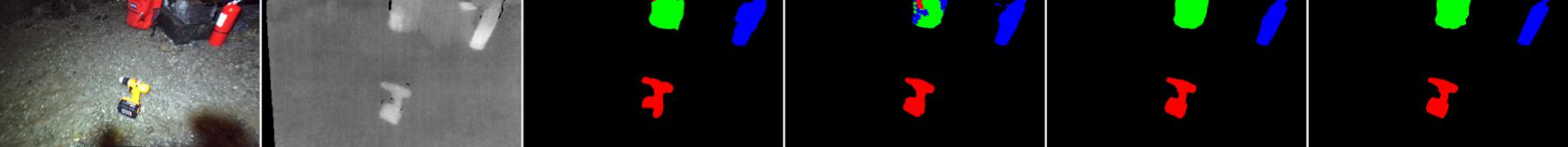}} \vspace{-0.02in} \\
\multicolumn{6}{c}{\includegraphics[width=0.98\linewidth]{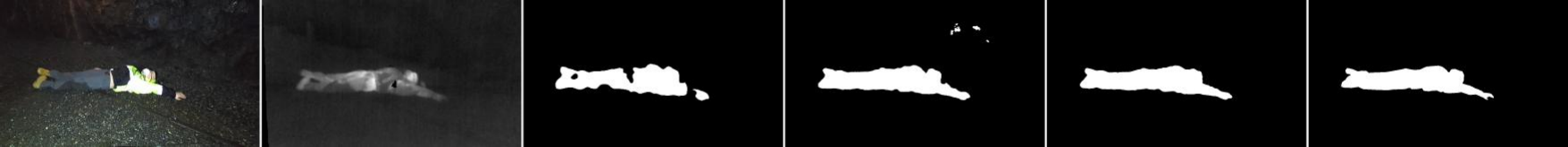}} \vspace{-0.02in} \\
{\footnotesize (a) RGB image} & {\footnotesize (b) THR image} & {\footnotesize (c) PSTNet~\cite{shivakumar2019pst900} } & {\footnotesize (d) CMXNet~\cite{liu2022cmx}} &  {\footnotesize (e) Ours (Swin-B)} &  {\footnotesize (e) GT}  \\ 
\multicolumn{6}{c}{\includegraphics[width=0.98\linewidth]{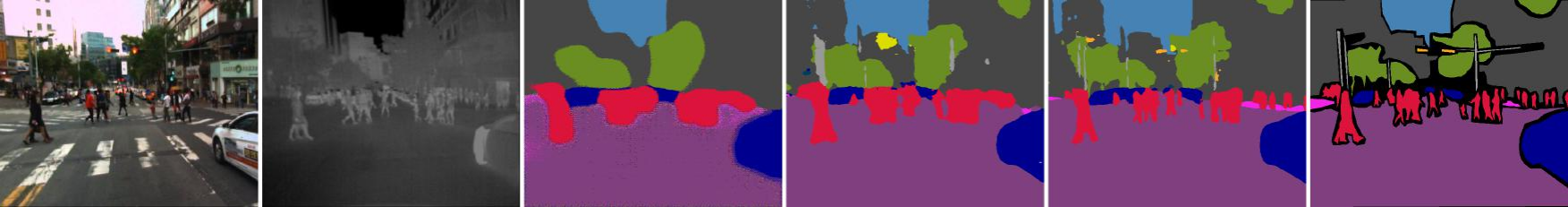}} \vspace{-0.02in} \\
\multicolumn{6}{c}{\includegraphics[width=0.98\linewidth]{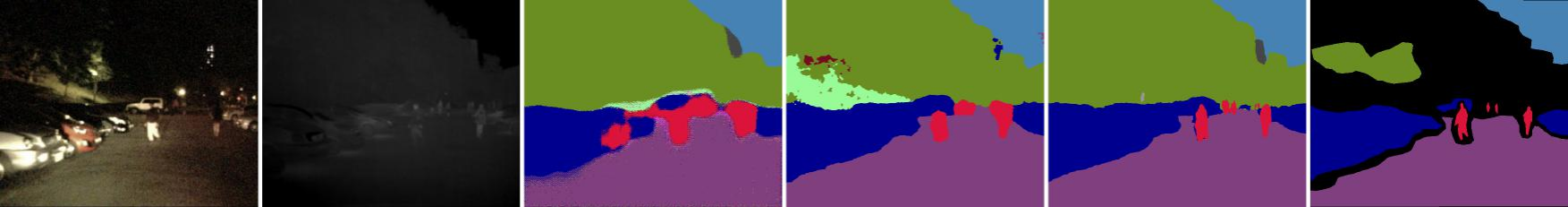}} \vspace{-0.02in} \\ 
\includegraphics[width=0.160\linewidth]{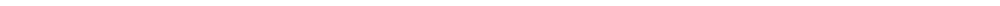} & \includegraphics[width=0.160\linewidth]{Images/images/dummy.png} & 
\includegraphics[width=0.160\linewidth]{Images/images/dummy.png} & \includegraphics[width=0.160\linewidth]{Images/images/dummy.png} & 
\includegraphics[width=0.160\linewidth]{Images/images/dummy.png} & \includegraphics[width=0.160\linewidth]{Images/images/dummy.png} \vspace{-0.1in} \\ 
{\footnotesize (a) RGB image} & {\footnotesize (b) THR image} & {\footnotesize (c) RTFNet~\cite{sun2019rtfnet} } & {\footnotesize (d) CMXNet~\cite{liu2022cmx}} &  {\footnotesize (e) Ours (Swin-B)} &  {\footnotesize (e) GT}  \\ 
\end{tabular}
}
\end{center}
\vspace{-0.1in}
\caption{{\bf Qualitative comparison for semantic segmentation of RGB-T images on MFNet~\cite{ha2017mfnet}, PST900~\cite{shivakumar2019pst900}, and KP~\cite{hwang2015multispectral} datasets.} 
The first two rows are qualitative comparisons of MFNet dataset, the next two rows are PST 900 dataset results, and the remaining rows are KP dataset results. 
The proposed method shows reliable and accurate segmentation results across all datasets, including day-light, low-light, noisy images, and harsh cave conditions. Further results can be found in the supplementary video.}
\label{fig:qualitative_results}
\vspace{-0.1in}
\end{figure*}

\subsection{RGB-T Semantic Segmentation Datasets} 
In this study, we employ three publicly available RGB-T datasets to train and evaluate the proposed method.

\textbf{MFNet dataset~\cite{ha2017mfnet}} consists of 820 daytime and 749 nighttime RGB-thermal images of urban driving scenes with a resolution of 640$\times$480. 
The dataset provides semantic labels for nine classes, including one unlabeled class and eight classes of common objects.

\textbf{PST900 dataset~\cite{shivakumar2019pst900}} provides 894 RGB-thermal images with a resolution of 1280$\times$720, taken under the cave and subterranean environments for DARPA Subterranean Challenge.
The dataset contains annotated segmentation labels for five classes, including one background class (\ie, unlabeled) and four object classes.

\textbf{KP dataset~\cite{hwang2015multispectral}} is RGB-T paired urban driving scene dataset, providing 95K video frames (62.5K for daytime and 32.5K for nighttime) with a resolution of 640$\times$512.
Originally, KAIST Multispectral Pedestrian Detection (KP) dataset provides detection bounding box labels only.
But, thankfully, Kim~\etal~\cite{kim2021ms} provides annotated semantic segmentation labels for 503 daytime and 447 nighttime images.
The labels include 19 object classes, which are the same classes as Cityscapes~\cite{cordts2016cityscapes} dataset.

However, the dataset splits of Kim~\etal~\cite{kim2021ms} are undesirable to common RGB-T semantic segmentation network training.
We divided 950 training images into 499 for training, 140 for validation, and 311 for testing. 
Daytime and nighttime images were appropriately distributed in each set.
We provide our train/val/test splits that were used to train our network and other networks on KP dataset.

\subsection{Implementation Details}
\label{subsec:impl_detail}
\textbf{Mask2Former.} We employ the Swin transforemr~\cite{liu2021swin} (tiny, small, and base) as our backbone model.
We use the multi-scale deformable attention Transformer (MSDeformAttn)~\cite{zhu2020deformable} as a pixel decoder.
We adopt the same Transformer decoder with DETR~\cite{carion2020end} for the Transformer decoder. 
The query $N$ is set to 100 by default.

\textbf{Training Settings.} 
Our proposed method is implemented with PyTorch~\cite{paszke2019pytorch} and Detectron2~\cite{wu2019detectron2} library on a machine equipped with two NVIDIA RTX A6000 GPUs. 
The following training setting is used for all datasets.
We use AdamW optimizer~\cite{loshchilovdecoupled} and poly learning rate scheduler~\cite{chen2017deeplab} with an initial learning rate of $10^{-4}$.
We train all segmentation networks with a batch size of 14 for 35K iterations.
We utilize Swin transformer model~\cite{liu2021swin} pre-trained on ImageNet-1K~\cite{russakovsky2015imagenet} (\ie, tiny(T), small(S), and base(B)) as a backbone model. 
We apply random color jittering~\cite{liu2016ssd}, random horizontal flipping, and random cropping to RGB and thermal images as data augmentation.
For the coefficients of loss functions, we set $\lambda_{cls}$ as 2.0 for predictions matched with a GT label and as 0.1 for the "no object" (\ie, no match with any GT labels) by following~\cite{cheng2022masked}.
Also, the coefficient $\lambda_{ce}$ and $\lambda_{dice}$ are set to 5.0

\subsection{RGB-Thermal Semantic Segmentation}
\label{subsec:rgb_thermal_semantic_segmentation}
In this section, we compare our proposed method with the previous RGB-T semantic segmentation networks~\cite{ha2017mfnet,sun2019rtfnet,shivakumar2019pst900,sun2020fuseseg,zhou2021gmnet,xu2021attention,zhang2021abmdrnet,deng2021feanet,liu2022cmx} on three benchmarks.
We use mean Intersection-over-Union (mIoU) to evaluate the performance of semantic segmentation.

\subsubsection{Evaluation on MFNet Day-Night Dataset~\cite{ha2017mfnet}}
\label{subsec:eval_mf_day_night}
The quantitative and qualitative comparison results are shown in~\tabref{tab:Exp_segmentation_total}-(a) and~\figref{fig:qualitative_results}.
We trained the RGB-T Mask2Former~\cite{cheng2022masked}, as described in~\secref{sec:method}, along with our proposed method.
Also, we provide a variant of the network with Swin-T, Swin-S, and Swin-B backbone models.
Compared to the previous state-of-the-art method (\ie, CMXNet~\cite{liu2022cmx}), our approach leads to a 1.7\% performance gain in the mIoU metric.
Furthermore, our methods (Swin-S and B) achieve the best or second-best performance in most IoU metrics over nine classes.

\subsubsection{Evaluation on PST900 Dataset~\cite{shivakumar2019pst900}}
\label{subsec:eval_pst_day_night}
For the PST900 benchmark, our model (\ie, Swin-B) achieves a high performance improvement by 3.9\% against previous state-of-the-art result (\ie, GMNet~\cite{zhou2021gmnet}), as shown in~\tabref{tab:Exp_segmentation_total}-(b).

\subsubsection{Evaluation on KP Day-Night Dataset~\cite{hwang2015multispectral,kim2021ms}}
\label{subsubsec:eval_kp_dataset}
KP dataset has a more diverse and numerous number of classes compared to the MF~\cite{ha2017mfnet} and PST900~\cite{shivakumar2019pst900} datasets.
Therefore, the increased complexity in the dataset makes it more difficult to accurately segment the objects in the RGB-T images, requiring more advanced techniques and consideration of multi-modality inputs. 
We trained publicly available RGB-T semantic segmentation networks (\ie, MFNet~\cite{ha2017mfnet}, RTFNet~\cite{sun2019rtfnet}, CMXNet~\cite{liu2022cmx}) on the KP dataset with their provided code bases.

As shown in~\tabref{tab:Exp_segmentation_total}-(c), our method (\ie, Swin-B) achieves 9.0\% performance improvement against CMXNet~\cite{liu2022cmx}.
Also, \figref{fig:qualitative_results} shows that our method shows precise and accurate segmentation quality in partially occluded, noisy, and cluttered environments.
This implies that the proposed complementary random masking and self-distillation loss make the network learn to extract non-local and complementary representations from each modality, even in challenging conditions. 
We think the results show that as the complexity of semantic segmentation is higher, our proposed method is more helpful in achieving accurate and robust semantic segmentation performance from RGB and thermal images.





\begin{table}[t]
\caption{\textbf{Ablation study on MFNet dataset~\cite{ha2017mfnet}.} 
We conduct an ablation study of the proposed method and various masking strategies. Swin-S backbone is used for ablation study.
}
\vspace{-0.1in}
\begin{center}
\subfloat[Ablation study of loss functions] 
{
    \resizebox{0.90\linewidth}{!}
    {
        \def\arraystretch{1.1}
        \footnotesize
        \begin{tabular}{c|cccc|c}
        \toprule
        Method & $L_{MWS}$ & $CRM$ & $L_{SDC}$ & $L_{SDN}$ & {\bf mIoU}\\ 
        \hline \hline
        Baseline & & & 	&		       & 59.5 \\ \hline
         (1)    & \checkmark & & &               				    & 60.3 \\
         (2)    & \checkmark & \checkmark & 			  &             & 60.7 \\ 
         (3)    & \checkmark & \checkmark & \checkmark & 			& 61.0 \\ 
         (4)    & \checkmark & \checkmark &            & \checkmark	& 60.9 \\ \hline
         Ours   & \checkmark & \checkmark & \checkmark & \checkmark	& \textbf{61.2} \\
        \bottomrule
        \end{tabular}
        \label{tab:abl_loss}
    }
}

\subfloat[Ablation study of complementary masking strategies] 
{
    \resizebox{0.95\linewidth}{!}
    {
        \def\arraystretch{1.5}
        \footnotesize
        \begin{tabular}{c|ccccc}
        \toprule
        Method & Square & Patch (8) & Patch (16) & Patch (32) & Patch (64) \\ 
        \hline \hline
        {\bf mIoU} & 59.7 & 59.8 & 60.8 & \textbf{61.2} & 61.1 \\
        \bottomrule
        \end{tabular}
        \label{tab:abl_crm}
    }
}
\end{center}
\vspace{-0.2in}
\label{tab: ablation study}
\end{table}


\subsection{Ablation Study}
\label{sec:ablation_study}


In this ablation study, we investigate the components of the proposed method, as shown in~\tabref{tab: ablation study}-(a).
Baseline indicates an RGB-T Mask2Former model that is modified to take RGB and thermal image inputs, as described in~\secref{sec:method}.
Our empirical finding demonstrates that modality-wise supervision loss $L_{MWS}$, which provides supervision for each prediction result from multiple input modalities, yields +0.8\% performance gain compared to a single supervised loss for RGB-thermal pairs (\ie, Baseline).

Also, applying complementary random masking $CRM$ brings +0.4\% performance improvement by pushing the network to segment and classify objects even when partially occluded inputs are provided.
The self-distillation losses for complementary and non-local representation ($L_{SDC}$ and $L_{SDN}$) are getting +0.3\% and +0.2\% improvement, respectively.
$L_{SDC}$ aims to make the network learn to extract complement representation when one modality information is missing.
$L_{SDN}$ aims to make each modality extract robust representation to segment objects based on their non-local context rather than local features. 
Lastly, when all components are combined together, we get +1.7\% performance improvement compared to the Baseline model.

\begin{figure*}[t]
\begin{center}
{
\begin{tabular}{c@{\hskip 0.005\linewidth}c@{\hskip 0.005\linewidth}c@{\hskip 0.005\linewidth}c@{\hskip 0.005\linewidth}c@{\hskip 0.005\linewidth}c}
\includegraphics[width=0.16\linewidth]{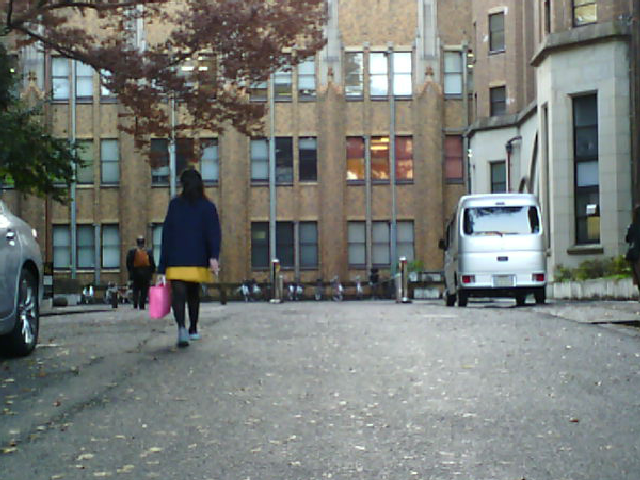} &
\includegraphics[width=0.16\linewidth]{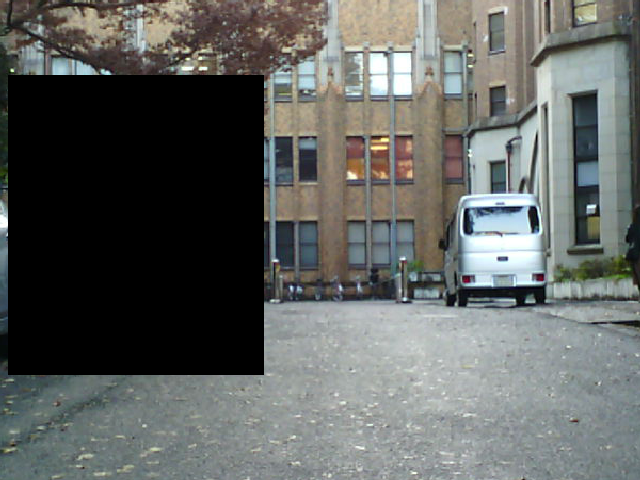} &
\includegraphics[width=0.16\linewidth]{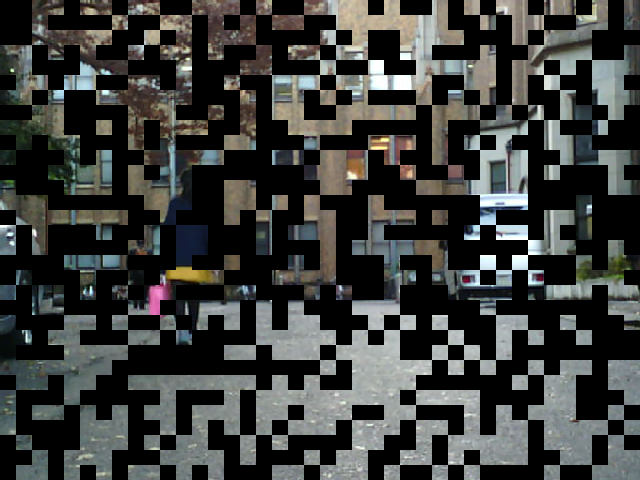} &
\includegraphics[width=0.16\linewidth]{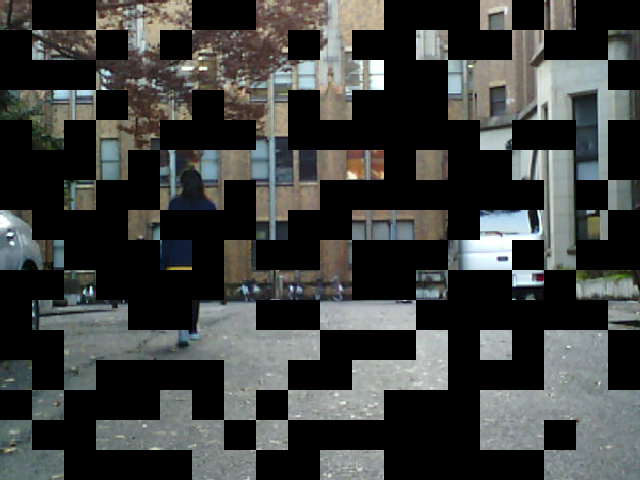} &
\includegraphics[width=0.16\linewidth]{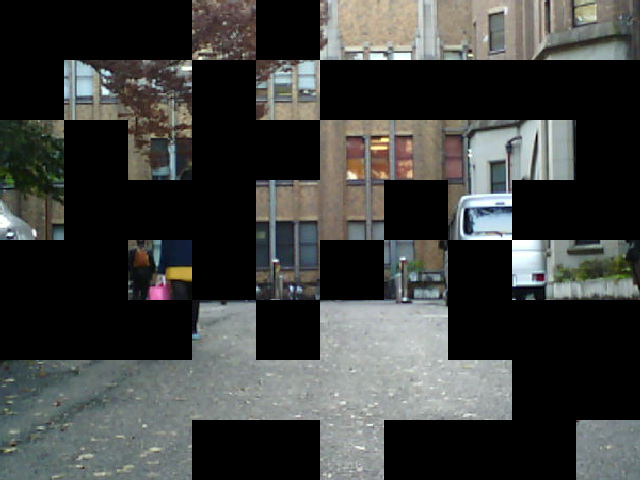} &
\includegraphics[width=0.16\linewidth]{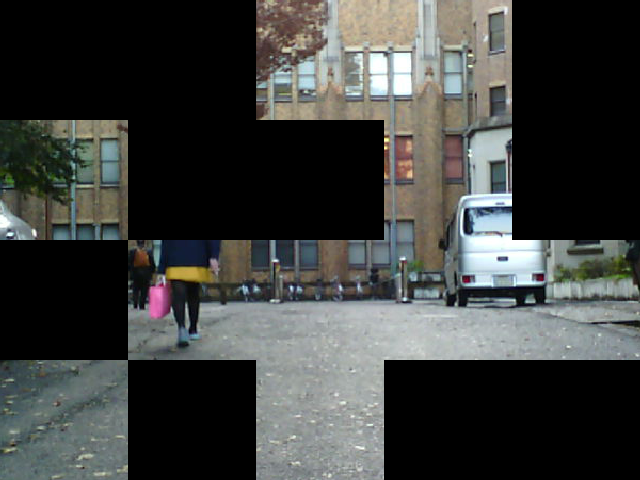} \\
\includegraphics[width=0.16\linewidth]{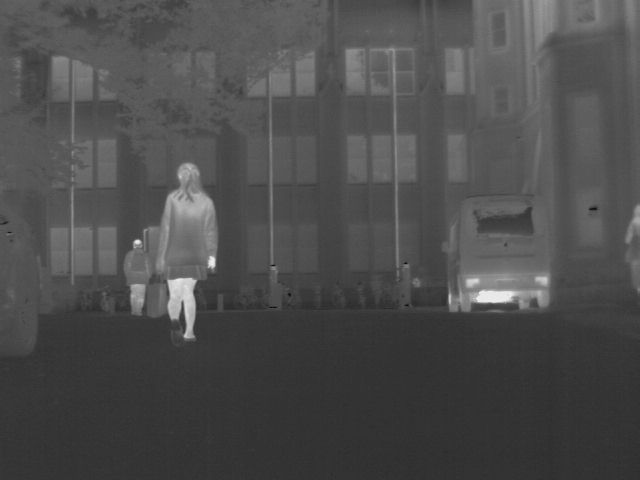} &
\includegraphics[width=0.16\linewidth]{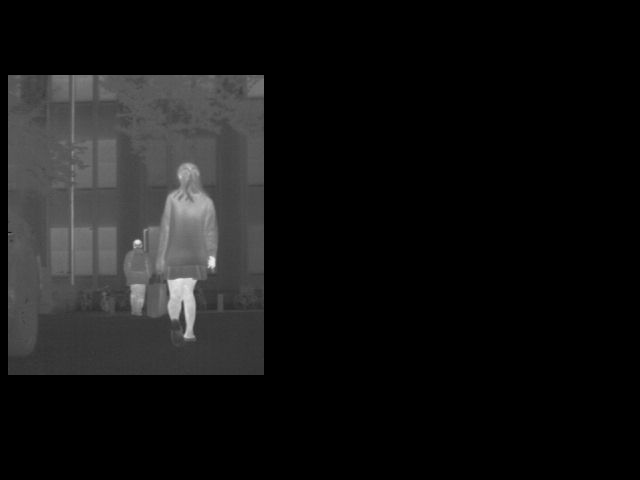} &
\includegraphics[width=0.16\linewidth]{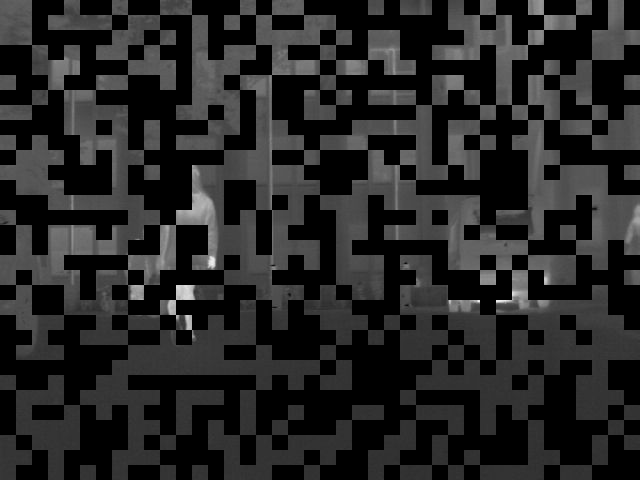} &
\includegraphics[width=0.16\linewidth]{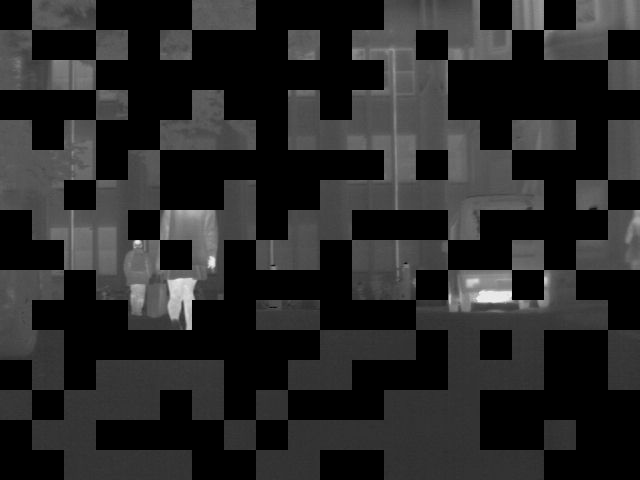} &
\includegraphics[width=0.16\linewidth]{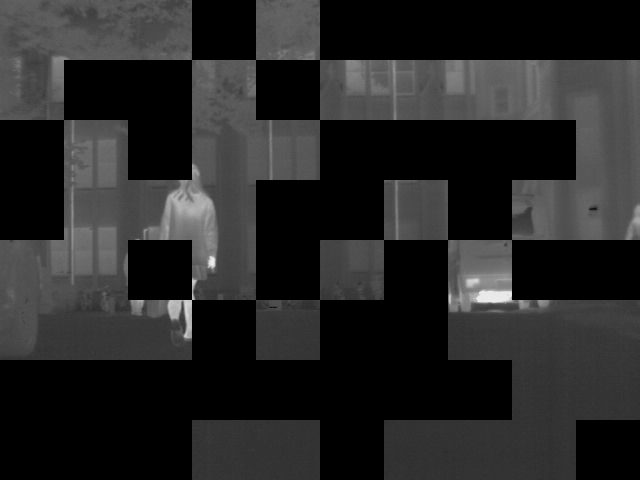} &
\includegraphics[width=0.16\linewidth]{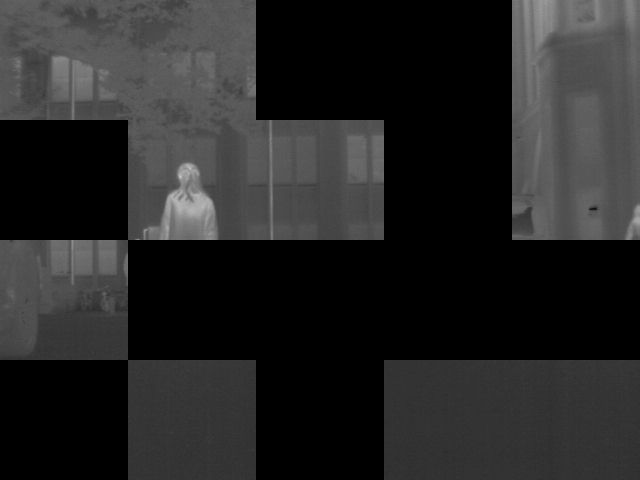} \\
{\footnotesize (a) RGB/THR images} & {\footnotesize (b) Random square } & {\footnotesize (c) Random patch (8)} & {\footnotesize (d) Random patch (16)} & {\footnotesize (e) Random patch (32)} & {\footnotesize (f) Random patch (64)} \\
\end{tabular}

}
\end{center}
\vspace{-0.1in}
\caption{{\bf Illustration of various complementary masking strategies.} 
Square masking randomly masks a square area with half the height and width of the image in a random position.
Patch masking randomly masks half an image (\ie, 0.5 ratios) with patches of different sizes (\eg, 8, 16, 32, 64).
}
\label{fig:ablation}
\vspace{-0.2in}
\end{figure*}

\subsubsection{Complementary Random Masking} 
We study various types of complementary masking strategies, as shown in~\figref{fig:ablation} and~\tabref{tab: ablation study}-(b).
Square masking randomly masks a square area with half the height and width of the image in a random position.
Patch masking randomly masks half an image (\ie, 0.5 ratios) with patches of different sizes (\eg, 8, 16, 32, 64).

Generally, complementary random masking shows better performance than the Baseline model. 
But, the patch size or masking scheme seems to be importnat for network performance.
For example, the complementary random square masking may hinder learning non-local representation from $L_{SDN}$ loss by masking out a wide area.
Similarly, too tiny patch size is also undesirable to learn complementary representations for each modality.
Generally, random patches over a certain size show higher performance. 
Empirically, we found that patch size 32 shows the best performance. 

\section{Conclusion}
In this paper, we have proposed a complementary random masking strategy and self-distillation loss for robust and accurate RGB-Thermal semantic segmentation.
The proposed masking strategy prevents over-reliance on a single modality. 
It also improves the accuracy and robustness of the neural network by forcing the network to segment and classify objects even when one modality is partially available.
Also, the proposed self-distillation loss encourages the network to extract complementary and meaningful representations by enforcing class prediction consistency between clean and masked RGB-thermal pairs.
Based on the proposed method, we achieve state-of-the-art performance over three RGB-T semantic segmentation benchmarks.

\section*{ACKNOWLEDGMENT}
This work was supported by the Ministry of Trade, Industry and Energy (MOTIE) and Korea Institute of Advancement
of Technology (KIAT) through the International Cooperative R\&D program (P0019782). Also, this research was supported by a grant (P0026022) from R\&D Program funded by Ministry of Trade, Industry and Energy of Korean government.

{\small
\bibliographystyle{ieee_fullname}
\bibliography{egbib}
}

\section{Appendix}
In this supplementary material, we provide

\begin{itemize}
\item 
Experimental comparison with individual random masking.
\item 
Further qualitative results on MF~\cite{ha2017mfnet} dataset.
\item 
Further qualitative results on PST900~\cite{shivakumar2019pst900} dataset.
\item 
Further qualitative results on KP~\cite{hwang2015multispectral} dataset.
\end{itemize}

\section{Experimental Comparison with Individual Random Masking}
We investigate the importance of complementarity in the random masking strategy for RGB-T semantic segmentation.
A na\"ive masking strategy for multi-modal inputs is randomly masking each modality independently.
The experimental results for Individual Random Masking (IRM) are shown in~\tabref{tab:abl_irm}.
We conduct the experiments with our final model (\ie, $L_{MWS}$ + $CRM$ + $L_{SDC}$ + $L_{SDN}$). 
Here, we replaced the masking strategy with IRM instead of CRM.
We applied random masking to each modality for the IRM strategy, using a masking ratio between 0.3 and 0.7.

As shown in~\tabref{tab:abl_irm}, IRM shows poor performance in all cases.
We think individual masking can cause missing information regions in both modalities. 
Therefore, self-supervision (\eg, $L_{SDC}$, $L_{SDN}$) for these regions may provide undesirable training signals for the network. 
Also, this tendency gets severe when the masking ratio is high. 
On the other hand, our complementary random masking guarantees every region is valid and can be complemented at least from one modality.

\begin{table}[h]
\caption{\textbf{Comparison with Individual Random Masking (IRM) on MF dataset~\cite{ha2017mfnet}.} 
We compare our Complementary Random Masking method (CRM) with Individual Random Masking (IRM). Every experimental setup is identical except masking strategy. Swin-S backbone is used for the comparison.}
\vspace{-0.1in}
\begin{center}
\resizebox{0.95\linewidth}{!}
{
    \def\arraystretch{1.5}
    \footnotesize
    \begin{tabular}{c|c|ccccc}
    \toprule
    \multirow{2}{*}{Methods} & \multirow{2}{*}{\textbf{CRM}} & \multicolumn{5}{c}{\textbf{IRM (masking ratio)}} \\ \cline{3-7}
     &  & 0.3 & 0.4 & 0.5 & 0.6 & 0.7 \\ 
    \hline \hline 
    {\bf mIoU} & \textbf{61.2} & 60.4 & 60.2 & 59.7 & 59.7 & 59.6 \\
    \bottomrule
    \end{tabular}
}
\end{center}
\vspace{-0.1in}
\label{tab:abl_irm}
\end{table}

\section{Further Qualitative Results}
\begin{figure*}[t]
\begin{center}
{
\begin{tabular}{c@{\hskip 0.005\linewidth}c@{\hskip 0.005\linewidth}c@{\hskip 0.005\linewidth}c@{\hskip 0.005\linewidth}c@{\hskip 0.005\linewidth}c}
\multicolumn{6}{c}{\includegraphics[width=0.98\linewidth]{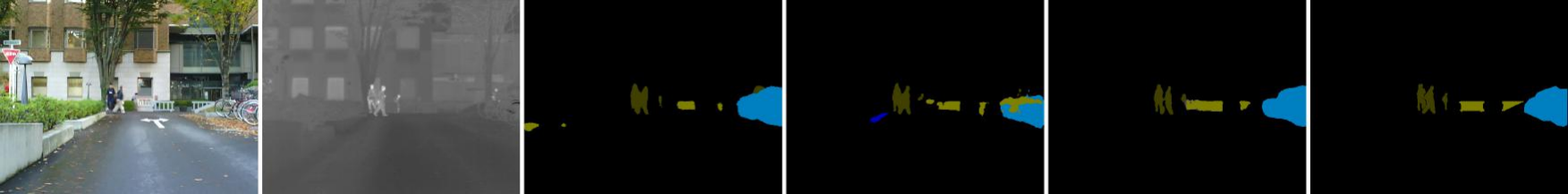}} \vspace{-0.02in} \\
\multicolumn{6}{c}{\includegraphics[width=0.98\linewidth]{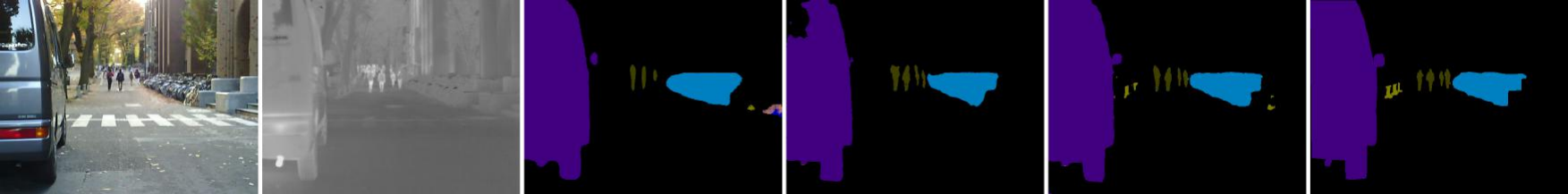}} \vspace{-0.02in} \\
\multicolumn{6}{c}{\includegraphics[width=0.98\linewidth]{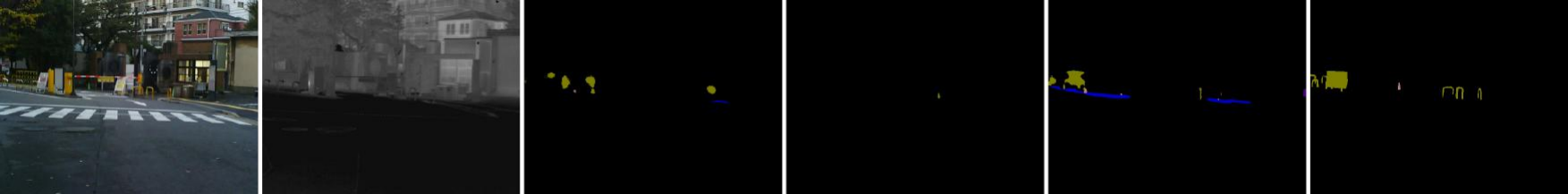}} \vspace{-0.02in} \\
\multicolumn{6}{c}{\includegraphics[width=0.98\linewidth]{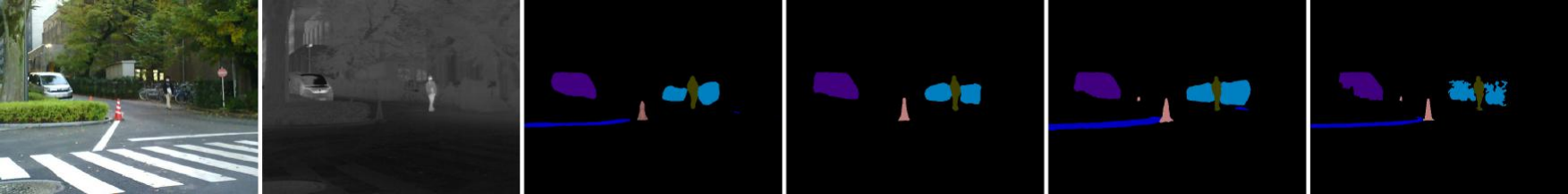}} \vspace{-0.02in} \\
\multicolumn{6}{c}{\includegraphics[width=0.98\linewidth]{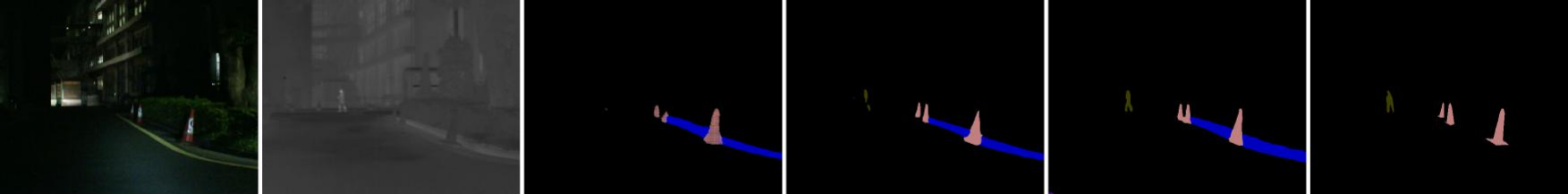}} \vspace{-0.02in} \\ 
\multicolumn{6}{c}{\includegraphics[width=0.98\linewidth]{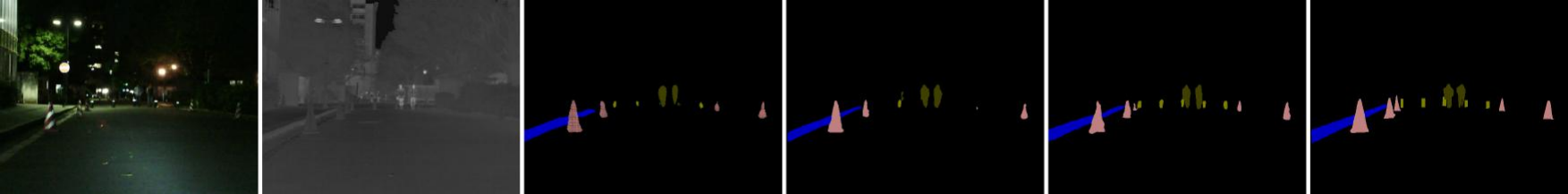}} \vspace{-0.02in} \\
\multicolumn{6}{c}{\includegraphics[width=0.98\linewidth]{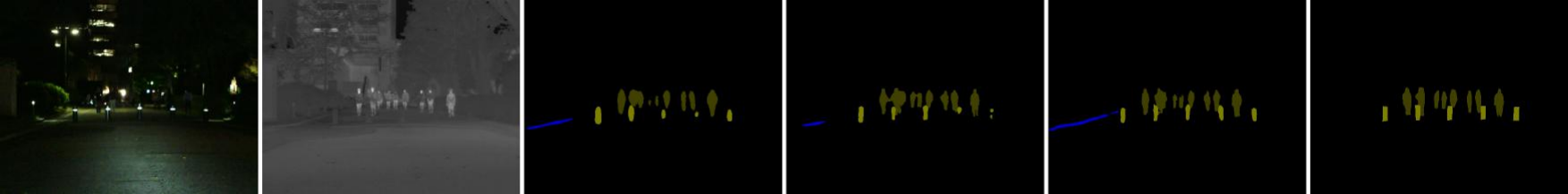}} \vspace{-0.02in} \\
\multicolumn{6}{c}{\includegraphics[width=0.98\linewidth]{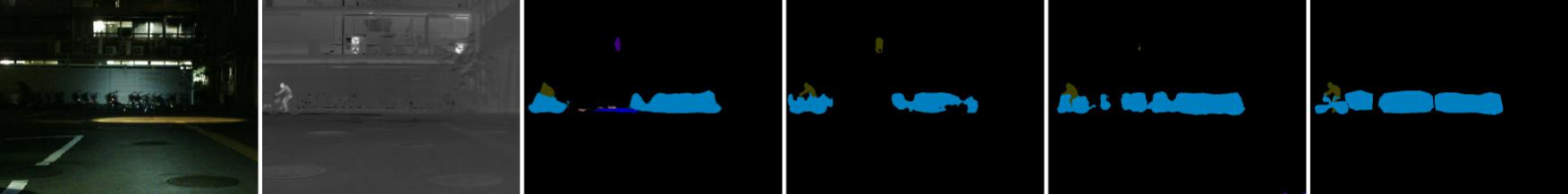}} \vspace{-0.02in} \\
\multicolumn{6}{c}{\includegraphics[width=0.98\linewidth]{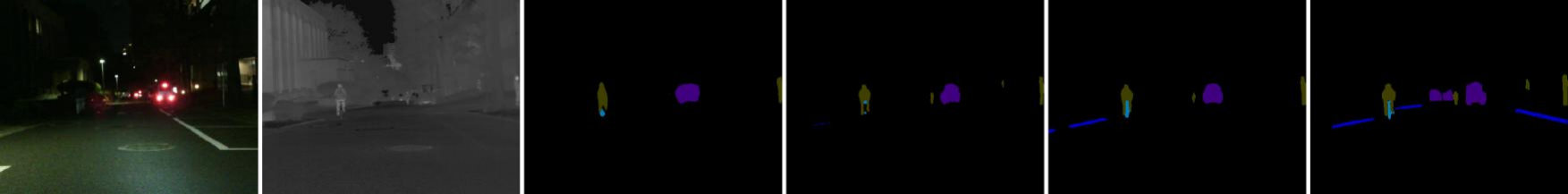}} \vspace{-0.02in} \\
\includegraphics[width=0.160\linewidth]{Images/images/dummy.png} & \includegraphics[width=0.160\linewidth]{Images/images/dummy.png} & 
\includegraphics[width=0.160\linewidth]{Images/images/dummy.png} & \includegraphics[width=0.160\linewidth]{Images/images/dummy.png} & 
\includegraphics[width=0.160\linewidth]{Images/images/dummy.png} & \includegraphics[width=0.160\linewidth]{Images/images/dummy.png} \vspace{-0.1in} \\ 
{\footnotesize (a) RGB image} & {\footnotesize (b) THR image} & {\footnotesize (c) RTFNet~\cite{sun2019rtfnet}} & {\footnotesize (d) CMXNet~\cite{liu2022cmx}} &  {\footnotesize (e) Ours (Swin-B)} &  {\footnotesize (e) GT}  \\ 
\end{tabular}
}
\end{center}
\vspace{-0.2in}
\caption{{\bf Qualitative comparison for semantic segmentation of RGB-T images on MF~\cite{ha2017mfnet} dataset.}}
\label{fig:supple_mf}
\vspace{-0.2in}
\end{figure*}

\begin{figure*}[t]
\begin{center}
{
\begin{tabular}{c@{\hskip 0.005\linewidth}c@{\hskip 0.005\linewidth}c@{\hskip 0.005\linewidth}c@{\hskip 0.005\linewidth}c@{\hskip 0.005\linewidth}c}
\multicolumn{6}{c}{\includegraphics[width=0.98\linewidth]{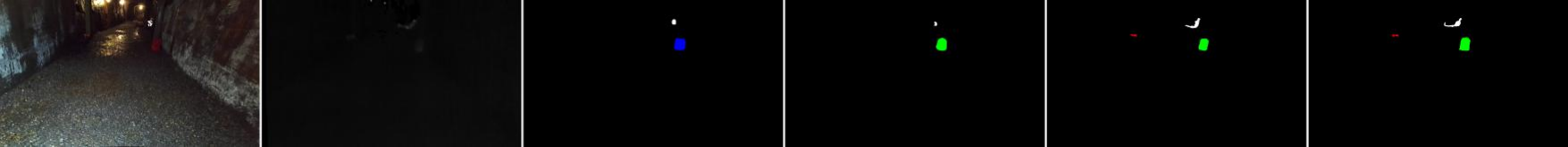}} \vspace{-0.02in} \\
\multicolumn{6}{c}{\includegraphics[width=0.98\linewidth]{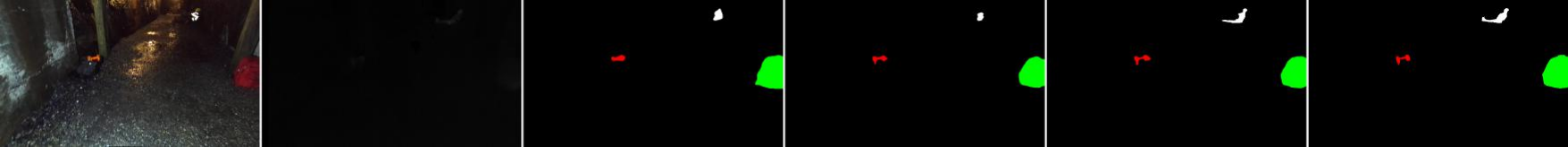}} \vspace{-0.02in} \\
\multicolumn{6}{c}{\includegraphics[width=0.98\linewidth]{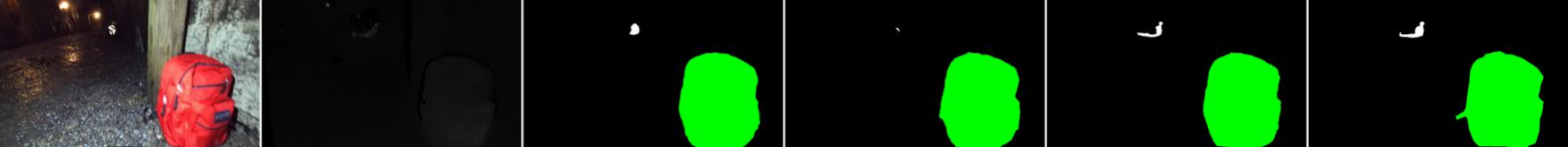}} \vspace{-0.02in} \\
\multicolumn{6}{c}{\includegraphics[width=0.98\linewidth]{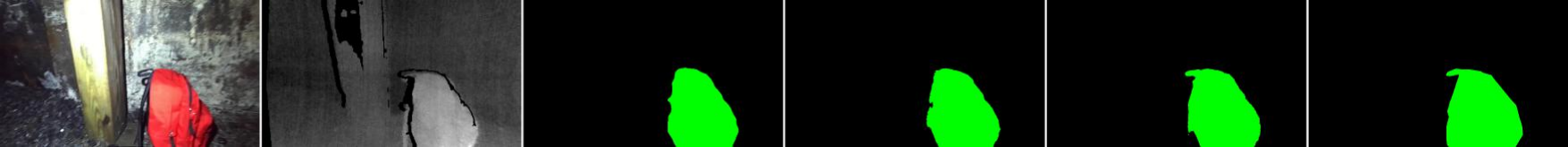}} \vspace{-0.02in} \\
\multicolumn{6}{c}{\includegraphics[width=0.98\linewidth]{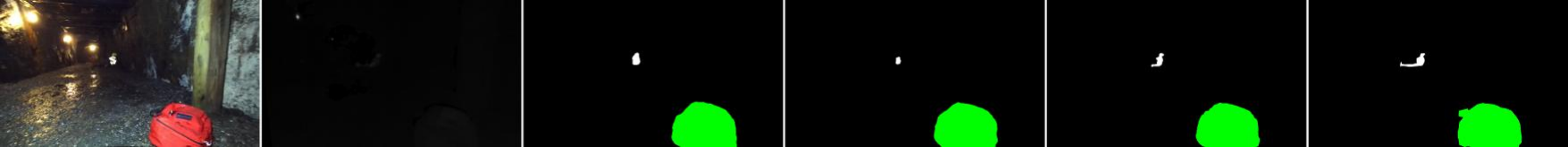}} \vspace{-0.02in} \\
\multicolumn{6}{c}{\includegraphics[width=0.98\linewidth]{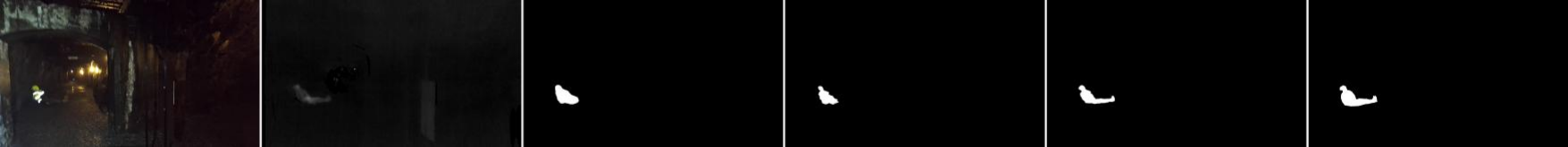}} \vspace{-0.02in} \\
\multicolumn{6}{c}{\includegraphics[width=0.98\linewidth]{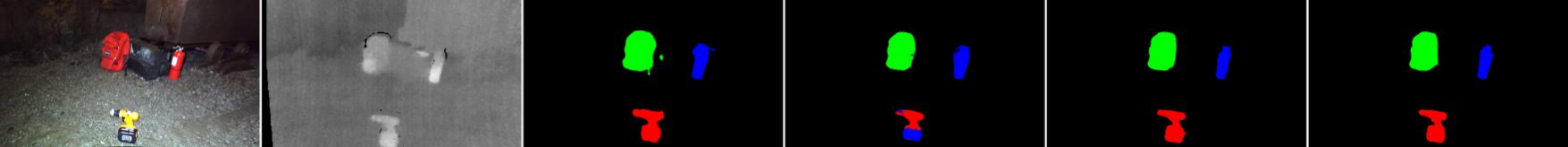}} \vspace{-0.02in} \\
\multicolumn{6}{c}{\includegraphics[width=0.98\linewidth]{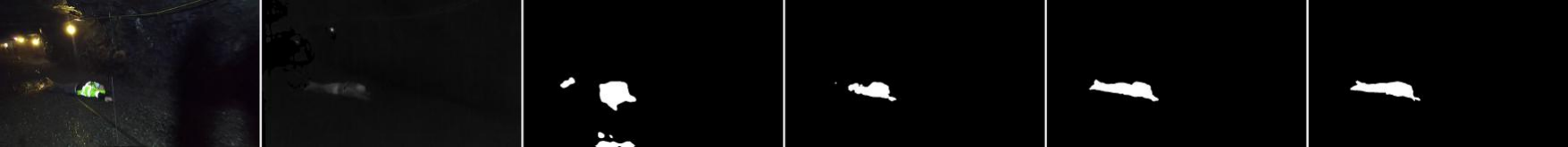}} \vspace{-0.02in} \\
\multicolumn{6}{c}{\includegraphics[width=0.98\linewidth]{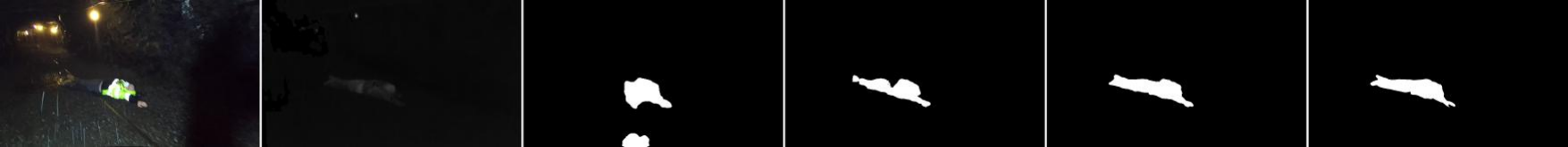}} \vspace{-0.02in} \\
\multicolumn{6}{c}{\includegraphics[width=0.98\linewidth]{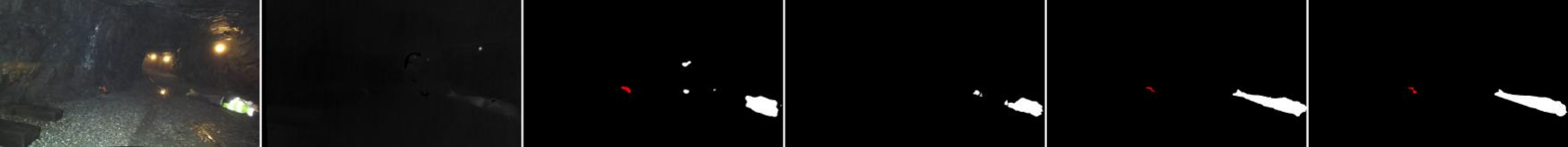}} \vspace{-0.02in} \\
\multicolumn{6}{c}{\includegraphics[width=0.98\linewidth]{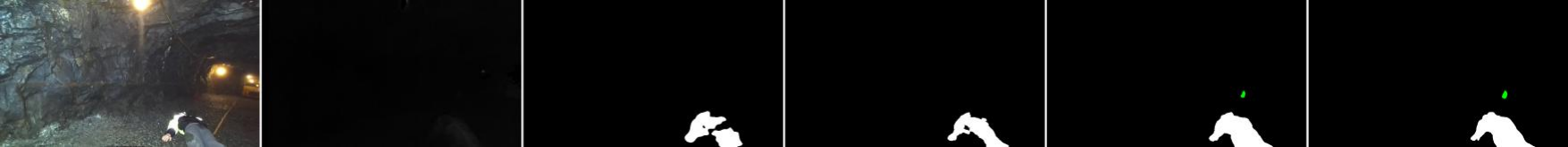}} \vspace{-0.02in} \\
\includegraphics[width=0.160\linewidth]{Images/images/dummy.png} & \includegraphics[width=0.160\linewidth]{Images/images/dummy.png} & 
\includegraphics[width=0.160\linewidth]{Images/images/dummy.png} & \includegraphics[width=0.160\linewidth]{Images/images/dummy.png} & 
\includegraphics[width=0.160\linewidth]{Images/images/dummy.png} & \includegraphics[width=0.160\linewidth]{Images/images/dummy.png} \vspace{-0.1in} \\ 
{\footnotesize (a) RGB image} & {\footnotesize (b) THR image} & {\footnotesize (c) PSTNet~\cite{shivakumar2019pst900} } & {\footnotesize (d) CMXNet~\cite{liu2022cmx}} &  {\footnotesize (e) Ours (Swin-B)} &  {\footnotesize (e) GT}  \\ 
\end{tabular}
}
\end{center}
\vspace{-0.2in}
\caption{{\bf Qualitative comparison for semantic segmentation of RGB-T images on PST900~\cite{shivakumar2019pst900} dataset.}}
\label{fig:supple_pst}
\vspace{-0.2in}
\end{figure*}

\begin{figure*}[t]
\begin{center}
{
\begin{tabular}{c@{\hskip 0.005\linewidth}c@{\hskip 0.005\linewidth}c@{\hskip 0.005\linewidth}c@{\hskip 0.005\linewidth}c@{\hskip 0.005\linewidth}c}
\multicolumn{6}{c}{\includegraphics[width=0.98\linewidth]{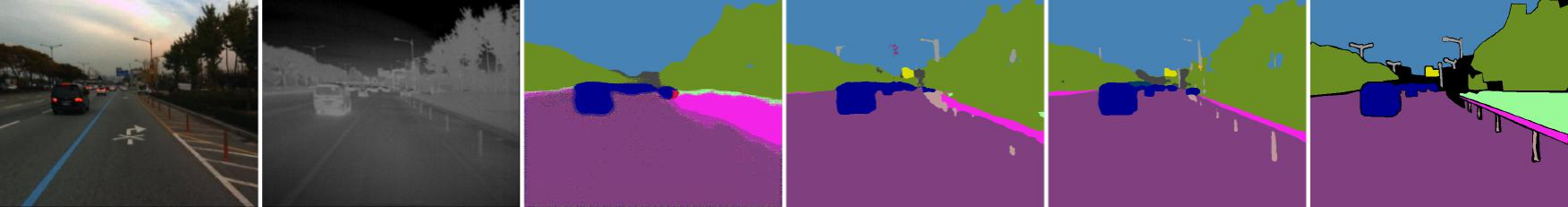}} \vspace{-0.02in} \\
\multicolumn{6}{c}{\includegraphics[width=0.98\linewidth]{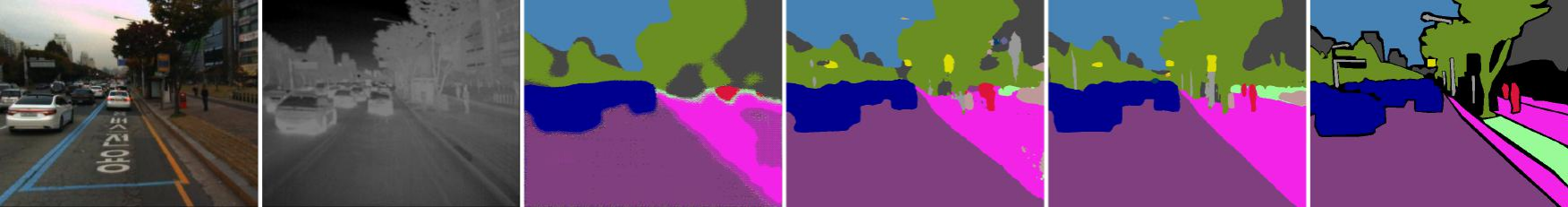}} \vspace{-0.02in} \\
\multicolumn{6}{c}{\includegraphics[width=0.98\linewidth]{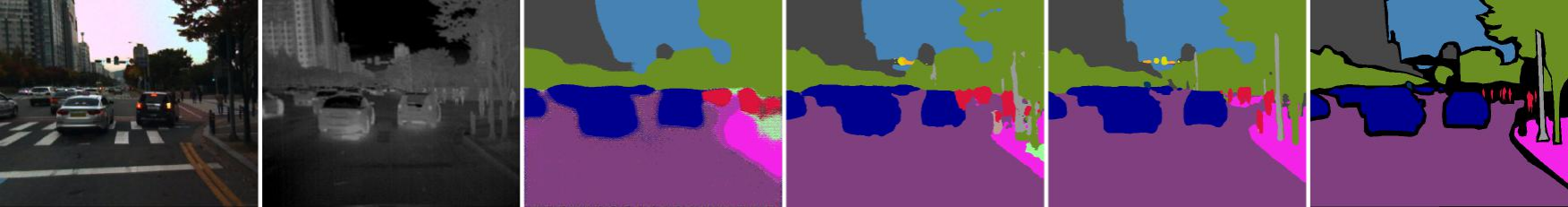}} \vspace{-0.02in} \\
\multicolumn{6}{c}{\includegraphics[width=0.98\linewidth]{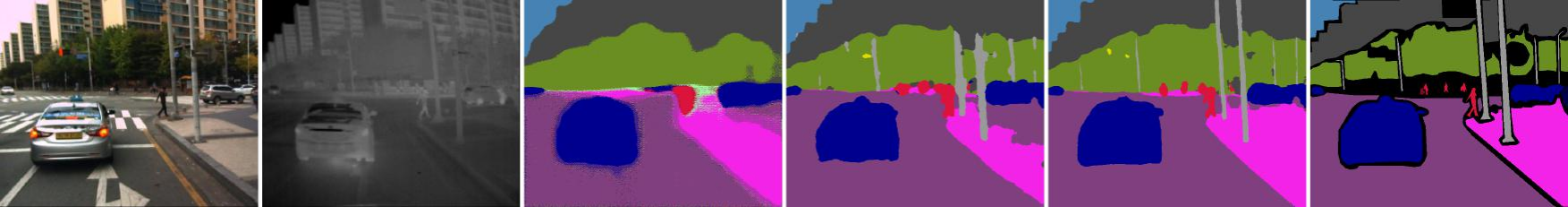}} \vspace{-0.02in} \\
\multicolumn{6}{c}{\includegraphics[width=0.98\linewidth]{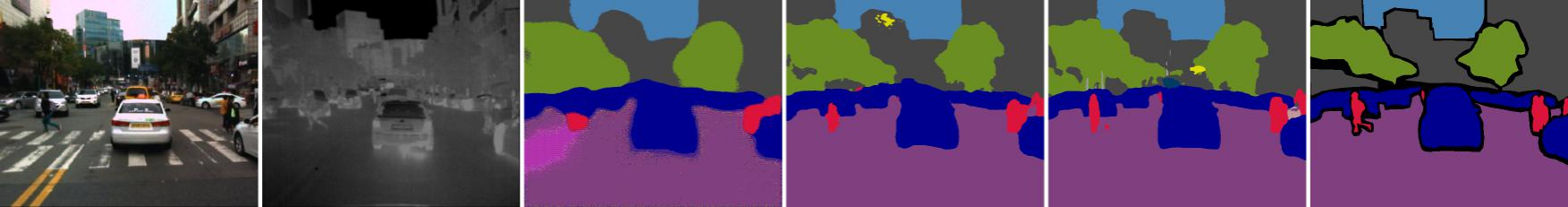}} \vspace{-0.02in} \\
\multicolumn{6}{c}{\includegraphics[width=0.98\linewidth]{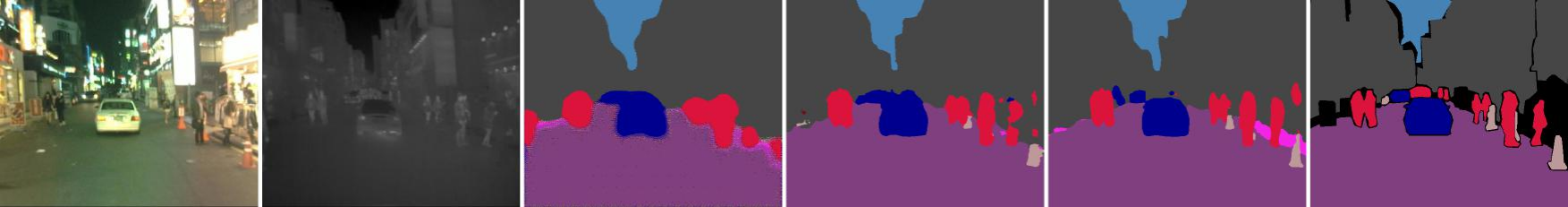}} \vspace{-0.02in} \\
\multicolumn{6}{c}{\includegraphics[width=0.98\linewidth]{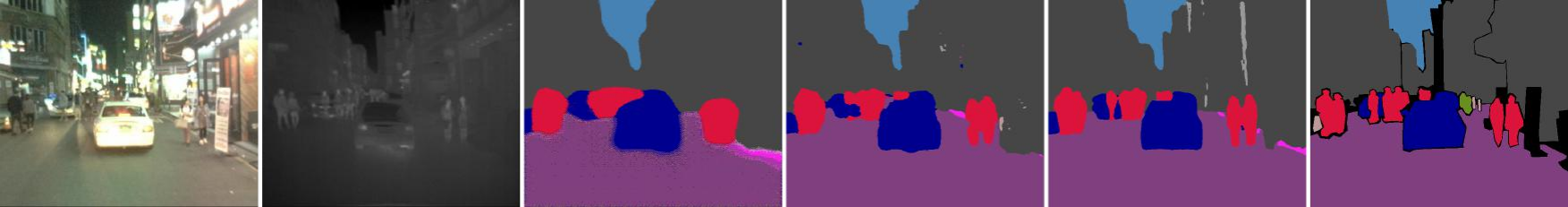}} \vspace{-0.02in} \\
\multicolumn{6}{c}{\includegraphics[width=0.98\linewidth]{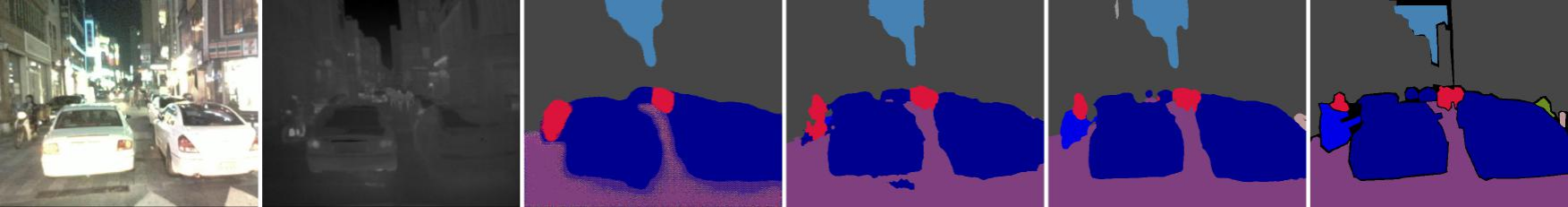}} \vspace{-0.02in} \\
\includegraphics[width=0.160\linewidth]{Images/images/dummy.png} & \includegraphics[width=0.160\linewidth]{Images/images/dummy.png} & 
\includegraphics[width=0.160\linewidth]{Images/images/dummy.png} & \includegraphics[width=0.160\linewidth]{Images/images/dummy.png} & 
\includegraphics[width=0.160\linewidth]{Images/images/dummy.png} & \includegraphics[width=0.160\linewidth]{Images/images/dummy.png} \vspace{-0.1in} \\ 
{\footnotesize (a) RGB image} & {\footnotesize (b) THR image} & {\footnotesize (c) RTFNet~\cite{sun2019rtfnet} } & {\footnotesize (d) CMXNet~\cite{liu2022cmx}} &  {\footnotesize (e) Ours (Swin-B)} &  {\footnotesize (e) GT}  \\ 
\end{tabular}
}
\end{center}
\vspace{-0.2in}
\caption{{\bf Qualitative comparison for semantic segmentation of RGB-T images on KP~\cite{hwang2015multispectral} dataset.}}
\label{fig:supple_kp}
\vspace{-0.2in}
\end{figure*}

We provide further qualitative results on MF~\cite{ha2017mfnet}, PST900~\cite{shivakumar2019pst900}, and KP~\cite{hwang2015multispectral} dataset, as shown in~\figref{fig:supple_mf},~\figref{fig:supple_pst}, and~\figref{fig:supple_kp}, respectively.

\end{document}